\documentclass[aps,reprint,showkeys]{revtex4} 
\usepackage{amsmath}
\usepackage{amssymb}
\usepackage{graphicx}%

\usepackage{array}
\usepackage{booktabs}
\usepackage{float}

\usepackage[usenames]{color}

\begin{document}

\title{Interpretable neural network system identification method for two families of second-order systems based on characteristic curves} 

\author{Federico J. Gonzalez}
\email{fgonzalez@ifir-conicet.gov.ar}
\author{Luis P. Lara}
\affiliation{Instituto de F\'{\i}sica Rosario (IFIR), CONICET-UNR, Bv.\ 27 de Febrero 210 bis, S2000EZP Rosario, Argentina}
\affiliation{Facultad de Ciencias Exactas, Ingenier\'ia y Agrimensura (UNR), Av. Pellegrini 250, S2000BTP Rosario, Argentina}

\begin{flushright}
\color{blue}
DOI: 10.1007/s11071-025-11744-6\
(Received 21 May 2025; accepted 23 August 2025; published 12 September 2025)
\end{flushright}


\begin{abstract}
Nonlinear system identification often involves a fundamental trade-off between interpretability and flexibility, often requiring the incorporation of physical constraints.  We propose a unified data-driven framework for second-order nonlinear systems that combines the mathematical structure of the governing differential equations with the flexibility of neural networks (NNs). At the core of our approach is the concept of characteristic curves (CCs), which represent individual nonlinear functions (e.g., friction and restoring components) of the system. Each CC is modeled by a dedicated NN, enabling a modular and interpretable representation of the system equation. 
We apply this formalism to two families of second-order systems (position- and velocity-dependent friction force models) that include both nonlinear friction and restoring forces. 
To demonstrate the versatility of the CC-based formalism, we introduce three identification strategies: (1) SINDy-CC, which extends the sparse regression approach of SINDy by incorporating the mathematical structure of the governing equations as constraints; (2) Poly-CC, which represents each CC using high-degree polynomials; and (3) NN-CC, which uses NNs without requiring prior assumptions about basis functions.
Our results show that all three approaches are well-suited for systems with simple polynomial nonlinearities, such as the van der Pol oscillator. 
In contrast, NN-CC demonstrates superior performance in modeling systems with complex nonlinearities and discontinuities, such as those observed in stick-slip systems.  The key contribution of this work is to demonstrate that the CC-based framework, particularly the NN-CC approach, can capture complex nonlinearities while maintaining interpretability through the explicit representation of the CCs. This balance makes it well-suited for modeling systems with discontinuities and complex nonlinearities that are challenging to assess using traditional polynomial or sparse regression methods, providing a powerful tool for nonlinear system identification.
\end{abstract}

\let\newpage\relax  
\maketitle

\section{Introduction}

System identification aims to infer mathematical models of dynamical systems from observed input-output data, enabling a precise determination of the system model which can then be used for forward simulations and control. 
Traditional methods, such as ARMAX (auto-regressive moving-average with exogenous inputs) for linear and NARMAX (nonlinear ARMAX) for nonlinear systems\cite{Ljung1999,Billings_2013} provided a foundational framework based on autoregressive modeling. 
In the last decade, data-driven techniques have significantly expanded the scope of identifiable systems. Among them, sparse regression techniques such as SINDy (sparse regression of nonlinear dynamical systems) have received attention for their ability to identify parsimonious models by selecting a small subset of terms from a user-defined library of candidate functions\cite{brunton2016,Egan2024}. Sparse regression methods offer interpretable models but require the user to define an appropriate basis set, which can be challenging in highly nonlinear or discontinuous systems. 

Other strategies include Gaussian process state-space models (GPSSMs)\cite{Fan2023}, which offer a probabilistic, nonparametric framework for modeling latent dynamics with uncertainty quantification, and Koopman operator-based methods\cite{Korda2016,Jin2025}, which approximate nonlinear systems using linear dynamics in a higher-dimensional space of observables. 
Standard GPSSMs and Koopman approaches are not principally designed for symbolic, human‑readable ordinary differential equation (ODE) discovery; they aim for flexible, accurate prediction and uncertainty quantification. Symbolic regression methods (e.g. PySR\cite{Cranmer2023PySR}) can discover functional forms from data without requiring a predefined model structure. By combining mathematical operators (e.g., $+$, $-$, $\times$, $\div$, $\sin$, $\log$, $\mathrm{abs}$), these methods return a Pareto front of candidate models balancing accuracy and complexity. While flexible and interpretable, symbolic regression is computationally intensive and often yields multiple equally valid solutions\cite{Schmidt2009,Makke2024}.

In parallel, machine-learning approaches have been applied to system identification both in a black‐box fashion and in physics‐informed frameworks. 
Physics-informed neural networks (PINNs) embed differential‐equation constraints into the training of a global network, offering a flexible way to incorporate prior physical knowledge while learning complex dynamics from data \cite{Sjoberg1994,Raissi2019,Faroughi2024,Brunton_Kutz_2022,Pilloneto2025}. Despite these advances, a common limitation persists: many methods either assume a specific set basis functions or produce models that are difficult to interpret in terms of underlying physical mechanisms\cite{Goodfellow2016,Ayankoso2023,Elaarabi2025}.

Recently, a method based on the concept of characteristic curves (CCs) has been proposed for identifying first-order nonlinear systems (the name is inspired by the I-V characteristic curves in electronics)\cite{Gonzalez2023,Gonzalez2024}. This method represents the system dynamics in terms of their nonlinear elements that intrinsically define the ordinary differential equation (ODE). For first-order systems, CCs have been identified using both Fourier-based analysis\cite{Gonzalez2023} and least-squares fitting based on polynomial expansions\cite{Gonzalez2024}.  
Although the Fourier-based analysis and its mathematical connection to polynomial expansions apply only to first-order systems, the broader concept of characteristic curves (CCs) can be extended to higher-order systems. In this work, we extend the CC concept for second-order systems, focusing on modular and interpretable modeling. While we present an implementation based on polynomial expansions (Poly-CC), trigonometric basis functions could also be considered. Although this direction is not pursued here, we note that it is not directly related to the theoretical framework developed in our previous works. 
Extending the CC-based framework to second-order systems also introduces new practical challenges, particularly in dealing with more general nonlinearities and discontinuities. 



To this aim, Sec.~\ref{sec:methodology} proposes a unified framework for the identification of second-order nonlinear systems using CCs. We explore three strategies: (i) Poly-CC, which approximates each CC using high-degree polynomials fitted via least-squares regression; (ii) SINDy-CC, which applies sparse regression to identify a minimal set of candidate functions and recover analytic expressions for each CC; and (iii) NN-CC, which models each CC using neural networks (NNs), without relying on predefined basis functions. 

SINDy-CC is a straightforward extension of the standard SINDy method that incorporates additional constraints to ensure the identified model strictly satisfies the underlying differential equation structures proposed in this work. By embedding these restrictions, SINDy-CC achieves improved accuracy and physical consistency compared to applying standard SINDy without such constraints, leading to more reliable system identification results. 

These three methods differ in terms of flexibility and extrapolation behavior. 
In Sec.~\ref{sec:results}, we evaluate the performance of each approach using two benchmark problems: the forced van der Pol oscillator and a stick–slip friction model. For the van der Pol system, all three methods recover the correct CCs, with SINDy-CC and Poly-CC achieving near-machine precision and NN-CC yielding comparable accuracy. For the stick–slip system, where the friction law is discontinuous, only NN-CC succeeds in capturing the correct dynamics, illustrating its flexibility.  We further discuss model validations and investigate extrapolation issues in Appendix A, where we introduce a simple linear extrapolation scheme to reduce extrapolation errors for the NN-CC method. Appendix B presents an additional test in which both model families are applied to a given dataset to determine which one is compatible with the data. 
Appendix C analyzes the impact of different NN architectures on the identification performance of the NN-CC method. Appendices D and E include additional test cases for various systems within the two considered families of second-order systems, using the NN-CC method.  The discussion and conclusion sections are presented in Secs.~\ref{sec:discussion} and \ref{sec:conclusions}, respectively.

\section{Methodology}\label{sec:methodology}




A wide range of second-order nonlinear dynamical systems can be categorized based on whether the friction force depends on position or velocity\cite{Warminski2019}. In this work, we consider two families of differential equations:


(i) Position-dependent friction models (a Liénard equation with external forcing), described by
\begin{equation}
\ddot{x}(t)+f_1(x(t))\,\dot{x}(t)+f_2(x(t))=F_{ext}(t), \label{eq:model:pos}%
\end{equation}
where x(t) is the dynamical variable, $f_1(x)$ represents a position-dependent friction force, $f_2(x)$ represents an elastic force, and $F_{ext}(t)$ is an external driving force (which is zero in the homogeneous case).

(ii) Velocity-dependent friction models (a generalized Rayleigh-type nonlinear oscillation with velocity-dependent friction and external forcing), described by
\begin{equation}
\ddot{x}(t)+f_3(\dot{x}(t))+f_4(x(t))=F_{ext}(t), \label{eq:model:veloc}%
\end{equation}
where $f_3(\dot{x})$ is the velocity-dependent friction force, and $f_4(x)$ is a position-dependent elastic force.



In the proposed CC-based framework, the nonlinear functions $f_{i}$ ($i=\{1,\dotsi,4\}$) are referred to as ``characteristic curves'' (CCs). 
Conceptually, these CCs constitute the fundamental building blocks of the methodology presented in this work. 
We propose three methods to parameterize and obtain these CCs, thereby enabling modeling of the underlying system, which can subsequently be used for forward simulations using the corresponding ODE. 
Conceptually, once the CCs are identified, the corresponding ODE can be directly integrated to predict the system response for other prescribed external forcing $F_{ext}(t)$ and initial conditions.  
In practice, model extrapolation occurs during forward simulations when system variables exceed the range covered by the training data, potentially resulting in inaccurate or unstable predictions. This work examines the consequences of such extrapolation and highlights how different CC-based approaches respond to this issue.

These CCs contribute to two primary force types: damping and elastic (restoring) forces. Damping forces are represented by the terms $f_1(x)\,\dot{x}$ and $f_3(\dot{x})$. While these terms usually represent dissipative forces (energy loss), they can also exhibit energy-injecting behavior (negative damping). Energy injection occurs when $f_1(x)<0$ or when $f_3(\dot{x})$ lies in quadrants II or IV of the $f_3(\dot{x})$ vs. $\dot{x}$ plane (also called active devices). 
In contrast, elastic or restoring forces are represented by the position-dependent terms $f_2(x)$ and $f_4(x)$, which account for the storage and release of potential energy (see analogies with capacitive and inductive elements in electronics in Refs.~\cite{Chua1980,Chua2000}).


Tables~\ref{tab:friction_position} and \ref{tab:friction_velocity} present a representative, though non exhaustive, set of physical systems whose governing equations share the same mathematical structure as those studied in this work (i.e., Eqs.~\ref{eq:model:pos} and \ref{eq:model:veloc}). 
These systems range from vibration analysis, stick-slip friction, and seismology (for modeling geological faults and earthquakes), to phonation (vocal fold oscillations), also including systems with turbulent drag, biomechanical muscle modeling\cite{Winters1987}, vehicle suspension systems\cite{Ozcan2013}, and fluid dynamics where damping depends nonlinearly on velocity.

Although we studied in detail the van der Pol oscillator and a stick-slip system, 
we also tested additional position- and velocity-dependent systems in Appendixes~D and E, respectively. While these examples highlight the broad applicability of the CC framework, further considerations, such as the inclusion of noise effects, may be required in real-world applications. These aspects are beyond the scope of this work and are left for future investigation.






\begin{table*}[]
\caption{ Representative second-order position-dependent friction nonlinear systems compatible with the CC formalism, based on the mathematical structure of Eq~\ref{eq:model:pos}. The nonlinear functions $f_1(x)$ and $f_2(x)$ in these systems can, in principle, be identified using the proposed method. Only a subset of these systems is explicitly tested in this work, while others are listed to illustrate the general applicability of the framework and to motivate future studies.}
\begin{tabular}{llll}
\hline \hline
 \multicolumn{4}{l}{Model 1:  $\ddot{x}+f_1(x)\, \dot{x}+f_2(x)=F_{ext}(t)$}    \\ 
 System & $f_1(x)$ &  $f_2(x)$ & $F_{ext}(t)$\\ \hline

\begin{tabular}[c]{@{}l@{}} Polynomial nonlinearities  \\ \;(a) Van der Pol Oscillator\cite{vanderPol1926,Farkas1994} \\\;(b) Nonlinear air damping\cite{Wu2020} \\\;(c) Duffing-Holmes oscillator\cite{Holmes1979,Burra2025}\\\;(d) Vocal fold oscillation\cite{Story1995,Cataldo2021,Zhao2023}\\\;(e) Shipping rolling\cite{Bikdash1994,Mahfouz2004,Takami2024}   \end{tabular} 
&
\begin{tabular}[c]{@{}l@{}}
$a_0+a_1\,x+a_2\,x^2+\dotsi$\;\; \\ $\mu(x^2-1)$ \\ $c_0+c_1\,x$ \\ $\delta+\beta\,x$ \\ $a+b\,x$ \\ $B_1+B_2\,x^2$
\end{tabular}
&
\begin{tabular}[c]{@{}l@{}}
$b_0+b_1\,x+b_2\,x^2+\dotsi$\;\; \\ $k\,x$ \\ $k\,x+a\,x^3$ \\ $a\,x+k\,x^3$ \\ $T_0\,x+k\,x^3$ \\ $\Delta\rho\,g\,x$
\end{tabular}
& 
\begin{tabular}[c]{@{}l@{}}
\\
$F_{ext}(t)$: Driven Force \\
$Q(t)$: Wind gusts/modal force \\
$F(t)$: Harmonic driven force\\
$P(t)$: Lung pressure \\
$F_{wave}(t)$: Wave forcing
\end{tabular}
\\ \vspace{-.2cm} \\ \vspace{.2cm} 
Electrostatic MEMS\cite{Chuang2010} & $c_e/(d-x)^3$ & $k\,x-\varepsilon_0\,A\,V^2/[2\,(d-x)^2]$ & V(t): AC voltage\\ \vspace{.1cm}
Impact oscillation\cite{Shaw1983,Shaw1985,Feng2018} & $c$ & 
\begin{tabular}[c]{@{}l@{}} $kx+k(x-x_0)_+$ \\ where $z_+=\max(z,0)$ \end{tabular} & 
F(t): Impact force  \\
\begin{tabular}[c]{@{}l@{}} Neural spiking\cite{Fitzhugh1961,Gerstner2002,Rudi2020} \\(FitzHugh-Naguma) \\ \end{tabular} & $\varepsilon\, b+1- x^2$ & $\varepsilon\, [ a+(1-b)\,x+b\,x^3/3]$ & I(t): input current\\

 \hline \hline
\end{tabular}
\label{tab:friction_position}
\end{table*}

\begin{table*}[]
\caption{Representative second-order velocity-dependent friction nonlinear systems compatible with the CC formalism, based on the mathematical structure of Eq~\ref{eq:model:veloc}. The nonlinear functions $f_3(\dot{x})$ and $f_4(x)$ in these systems can, in principle, be identified using the proposed method. Only a subset of these systems is explicitly tested in this work, while others are listed to illustrate the general applicability of the framework and to motivate future studies.
}

\begin{tabular}{llll}
 \hline \hline
 \multicolumn{4}{l}{Model 2:  $\ddot{x}+f_3(\dot{x})+f_4(x)=F_{ext}(t)$ }   \\ 
 System & $f_3(\dot{x})$ &  $f_4(x)$ & $F_{ext}(t)$\\ \hline
\begin{tabular}[c]{@{}l@{}} 
Polynomial nonlinearities\\
\;(a)  Duffing mass-spring-damper\cite{nayfeh1979nonlinear} \\\;(b) MEMS/NEMS Microresonators\cite{Senturia2001,Lifshitz2008,Younis2011}
\end{tabular}
& \begin{tabular}[c]{@{}l@{}} 
$a_0+a_1\,\dot{x}+a_2\,\dot{x}^2+\dotsi$ 
\\
$c\,\dot{x}+\dotsi$ 
\\
$c_1\,\dot{x}+c_3\,\dot{x}^3$
\end{tabular}
&
\begin{tabular}[c]{@{}l@{}} 
$b_0+b_1\,x+b_2\,x^2+\dotsi$
\\
 $k\, x+\alpha \,x^3$ \\
  $k_1\, x+k_3 \,x^3$
\end{tabular}
&
$F_{ext}(t)$: Driven Force 
\\ \vspace{-0.2cm} \\\vspace{0.2cm}
\begin{tabular}[c]{@{}l@{}} Dry friction oscillator (stick-slip)\cite{Olsson1998,Rill2023,GonzalezCarbajal2024}: \\  \;(a) Coulomb friction\cite{nayfeh1979nonlinear} \\ \;(b) Dieterich-Ruina (steady state)\cite{Dieterich1979part1,Dieterich1979part2} 
\\\;(c) Stribeck friction model\cite{Stribeck1902,CanudasdeWit1995}
\end{tabular} & \begin{tabular}[c]{@{}l@{}} \\ $c\,\dot{x}+\mu\,N\,\text{sign}(\dot{x})$ \\ $c\,\dot{x}+[p_0+a\ln(|\dot{x}|/v_0)]\,\text{sign}(\dot{x})$
\\ $c\,\dot{x}+[p_0+a\exp(-(|\dot{x}|/v_0)^b)]\,\text{sign}(\dot{x})$ \\


 \end{tabular}  & \begin{tabular}[c]{@{}l@{}} \\   $k\,x+\beta\,x^3+\dotsi$ \\ $k\,x+\dotsi $ \\ $k\,x+\dotsi$  \end{tabular} & $F_{ext}(t)$: Driven Force  \\ \vspace{-0.2cm} \\\vspace{0.2cm}
Damped pendulum\cite{Braza2020,Pal2023} & $a\,\dot{\theta}$ & $b \sin(\theta)$ &$\tau_{ext}(t)$: Driven torque\\
\begin{tabular}[c]{@{}l@{}} 
Piecewise spring\cite{Ji2004,Bustamante2025} 
\end{tabular} &  $c\;\dot{x}$ & $\displaystyle
    \begin{cases}
       k_1\,x  & ;\quad |x| < d  \\      
       k_2\,x  & ;\quad |x| \geq d  
    \end{cases}
$   & $F_{ext}(t)$: Driven force\\ \vspace{-0.2cm} \\\vspace{0.2cm}
Backlash/dead-zone models\cite{slotine1991,Nordin2002,karimov2021} & $c\,\dot{x}+b\,\text{sign}(\dot{x})$ (Coulomb friction) & 
$\displaystyle
    \begin{cases}
       k\,x + d  & ;\quad k\,x < -d  \\
       0         & ;\quad |k\,x| \leq d \\
       k\,x - d  & ;\quad k\,x > d  
    \end{cases}
$    
& $F_{ext}(t)$: Driven force \\
 \hline \hline
\end{tabular}
\label{tab:friction_velocity}
\end{table*}

The CC-based approach focuses on explicitly identifying the functions $f_1$ and $f_2$ for position-dependent systems and the functions $f_3$ and $f_4$ for velocity-dependent systems, using an input dataset defined by $x(t)$ and the driven force $F_{ext}(t)$. Specifically, the CC-approach  relies on representing the $f_{i}$ ($i=\{1,\dotsi,4\}$) functions on some general basis of functions (e.g., polynomials), or through implicitly defined functions (e.g., neural networks). These function definitions enables a unified methodology that is compatible with the mathematical structure of all the systems listed in Tables~\ref{tab:friction_position} and \ref{tab:friction_velocity}, without requiring system-specific adjustments.






In most applications, practitioners who are familiar with the mathematical structure of the system are typically confident about whether the real system follows a position- or velocity-dependent friction model. However, when this information is not available, both models can be tested in practice. Appendix~B shows that although both position- and velocity-friction models may yield low training errors for a given input dataset, only the correct model is capable of accurately reproducing the system dynamics when used in forward simulations. 
Furthermore, if neither model produces accurate forward simulations, this may indicate that the true system dynamics is not well captured by either equation, and a different family of ODEs may be required.




In this work, we implemented three parameterizations for both position- and velocity-dependent fiction models, thus obtaining a total of six code implementations. The three studied parameterizations are:

(i) Poly-CC: where each CC is approximated by a polynomial of degree $N$,
\begin{equation}
    f_i(z)\approx\sum_{j=0}^{N} c_{i,j}  \; z^j \; , 
\label{eq:polynomial:base}
\end{equation}
where $i=\{1,\dotsi,4\}$, $z =\{x, \dot{x}\}$, and $c_{i,j}$ are coefficients to be determined by least squares.

(ii) SINDy-CC: where each CC is represented by polynomials of degree $N$, similar to Poly-CC, but where coefficients are determined by sparse regression techniques.

(iii) NN-CC: where each CC is represented by one dedicated NN. Defining the output of a NN by a capital letter F, the CCs can be written as 
\begin{equation}
f_i(z)\approx \text{F}_i(z;\theta_i)\;,
    \label{eq:NN:base}
\end{equation}
where $i=\{1,\dotsi,4\}$, $z =\{x, \dot{x}\}$, and $\theta_i$ are the weight and bias parameters of the corresponding NN.


In the rest of this section, we give important details for each parametrization:
\begin{itemize}
    \item[(i)] 
In practice, high-degree polynomial expansions often suffer from numerical instability and poor conditioning. Two solutions can be used to address these problems: (a) regularization techniques (e.g., Tikhonov regularization\cite{Tikhonov1995,Freeden2018}, and L1 or L2 regularization\cite{Goodfellow2016}), applied in SINDy, and (b) domain rescaling or variable transformation, applied in Poly-CC. 
Hence, for practical implementation of Poly-CC, rather than fitting directly the expression in Eq.~\ref{eq:polynomial:base}, we introduced a shifted and normalized variable
\begin{equation}
    \hat{z}=\frac{z-A_0}{A_1} \;,
\end{equation}
where hyperparameters $A_0$ and $A_1$ are defined as
\begin{equation}
\begin{array}{cc}
A_0=&(\max(z)+\min(z))/2 \\ 
A_1=&(\max(z)-\min(z))/2  \; ,
\end{array}
\label{eq:polynomial:A0A1}
\end{equation}
so that $\hat{z}\in [-1,1]$ over the identification dataset. We then use 
\begin{equation}
    f_i(z)=\sum_{j=0}^{N} \hat{c}_{i,j} \left( \frac{z-A_0}{A_1}\right)^j \; , \; 
\label{eq:polynomial}
\end{equation}
for the system identification procedure. Once the coefficients $\{\hat{c}_{i,j}\}_{j=0}^N$ have been identified, the original coefficients $\{c_{i,j}\}_{j=0}^N$ in Eq.~\ref{eq:polynomial:base} can be recovered by using\cite{Gonzalez2024}
\begin{equation}
c_{i,j}=\hat{c}_{i,j}+\sum_{k=j+1}^N \binom{k}{j} \frac{(-A_0)^{k-j}}{A_1^k} \; \hat{c}_{i,k} \; ,
\label{eq:polynomial:coeffscj}
\end{equation}
which must be computed recursively from $j=N$ down to $j=0$. This rescaling strategy was shown to minimize coefficient scatter and improve fit robustness for first-order systems, greatly improving numerical stability for large $N$\cite{Gonzalez2024}.


\item[(ii)] SINDy-CC is a straightforward extension of the standard SINDy framework, adapted to fit within the CC approach. To this end, we incorporated constraints that ensure consistency with the mathematical structure of the two families of differential equations. Two separate implementations were developed, corresponding to  Eqs.~\ref{eq:model:pos} and \ref{eq:model:veloc}. 
For the practical implementation, we included the external force $F_{ext}$ as an input function ($u_0$). To enable a fair comparison with the other methods, we constrained the external input term to appear only as $c_0 u_0$ with $c_0 = 1$, thereby avoiding higher-order powers of $u_0$ in the identified equations. This constraint was implemented using the ``ConstrainedSR3'' module\cite{Kaptanoglu2021}. For the velocity-dependent model, we also disabled the flag ``include\_interaction'' to suppress cross terms between $x(t)$ and $\dot{x}(t)$, ensuring a correct identification of the corresponding ODE.  
Specifically, for the van der Pol system, the standard SINDy code (with external input) produced results identical to SINDy-CC, likely due to the simple polynomial structure of the system.
However, in the stick-slip system, the additional constraints were essential: without them, spurious mixed terms between $x(t)$ and $\dot{x}(t)$, as well as higher-order powers of $u_0$ appeared, leading to instabilities and divergence in forward simulations, resulting in errors exceeding $10^6$. Therefore, the standard SINDy approach was not included in this work.
In all the studied systems, SINDy-CC showed lower errors than the standard SINDy method. This is an expected outcome, given the inclusion of additional prior knowledge about the system.

%


\item[(iii)] For the implementation of the NN-CC method, we begin by analyzing the implementation of the position-dependent damping model (Eq.~\ref{eq:model:pos}). The unknown CCs $f_1(x)$ and $f_2(x)$ are represented by two independent feedforward NNs
\begin{equation}
    F_1(x;\theta_1) \; \text{and}\;      F_2(x;\theta_2) \; ,
\end{equation}
where $\theta_1$ and $\theta_2$ denote the respective weight and bias vector parameters. Given measurements of $x(t)$, $\dot{x}(t)$, $\ddot{x}(t)$ (where derivatives can be estimated from filtering procedures\cite{Kaptanoglu2021,Kaheman2022,Strebel2025}), along with the applied external forcing $F_{ext}(t)$, we define a training dataset $\{ x(t_i),\dot{x}(t_i),\ddot{x}(t_i),F_{ext}(t_i) \}_{i=0}^{N_{\text{data}}-1}$, where $N_{\text{data}}$ is the number of data points.  The NN-predicted forcing is then defined as 
\begin{equation}
\hat{F}_{ext}(t)=\ddot{x}(t)+F_1(x(t)\,;\,\theta_1)\; \dot{x}(t)+F_2(x(t)\,;\,\theta_2)\; , 
\label{eq:NN:predictedFext}%
\end{equation}
which recovers the true external forcing when $F_1$ and $F_2$ exactly reproduce the underlying CCs. 
To obtain $\theta_1$ and $\theta_2$, we minimize the mean-square error between the predicted and measured external forcings using the loss (or error) functional: 
\begin{equation}
    L(\theta_1,\theta_2)=\frac{1}{(t_{f}-t_0)} \int_{t_0}^{t_{f}} \left[\hat{F}_{ext}(t)-F_{ext}(t)\right]^2 dt \; ,
    \label{eq:loss}
\end{equation}
where $t_0$ and $t_f$ are the initial and final time data points ($t_f=t_{[N_{\text{data}}-1]}$). 
Optimization of $L$ with respect to both parameter sets $\{\theta_1,\theta_2\}$ yields the NN approximations of the CCs that best reproduce the dynamics under the given external forcing.


By substituting the expression for $F_{ext}(t)$ [Eq.~\ref{eq:NN:predictedFext}] into Eq.~\ref{eq:loss}, we determine the optimal parameter vectors $\theta_1^*$ and $\theta_2^*$ that minimize the discrepancy between predicted and observed forcing. Formally,
\begin{equation}
    \theta_1^*,\theta_2^*=\underset{\theta_1,\theta_2}{\text{argmin}} \; L(\theta_1,\theta_2) \; .
\end{equation}
Gradients of the loss with respect to the parameter vectors, $\nabla_{\theta_1}L$ and $\nabla_{\theta_2}L$, are computed via backpropagation, and parameters are updated simultaneously in each iteration using the gradient-based optimizer (e.g., Adam or stochastic gradient descent):
\begin{equation}
\begin{cases}
\theta_1 &\leftarrow \theta_1 -\eta\; \nabla_{\theta_1}L \;\;\; \\ \theta_2 &\leftarrow \theta_2 -\eta\; \nabla_{\theta_2}L \;, 
    \end{cases}
\end{equation}
where $\eta$ is the learning rate. 
With this methodology, weights and biases of both NNs are updated in tandem, guaranteeing that both NNs are trained on each iteration epoch.


Upon convergence, the networks $F_1$ and $F_2$ provide data-driven approximations of the true CCs. These networks may then be used for forward simulation under new initial conditions and/or external forcings using Eq.~\ref{eq:model:pos}. 
The overall workflow is depicted in Fig.~\ref{fig:NN_structure:pos}. An analogous procedure applies to the velocity-dependent friction model, which is depicted in Fig.~\ref{fig:NN_structure:vel}.

\begin{figure}[htpb]
    \centering
    \includegraphics[scale=0.12]{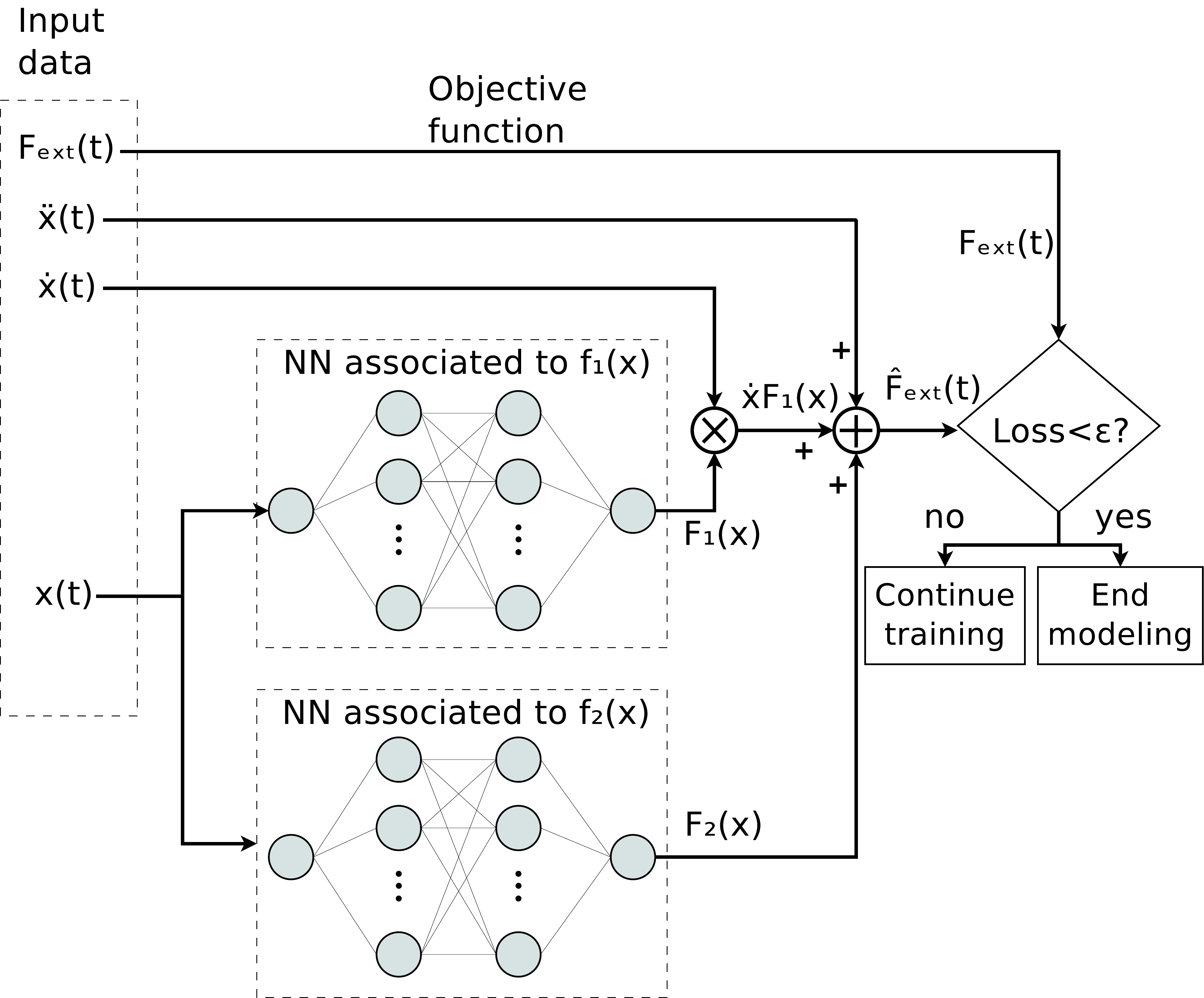}
    \caption{Schematic diagram for NN-CC implementation of the position-dependent friction model.}
    \label{fig:NN_structure:pos}
\end{figure}
\begin{figure}[htpb]
    \centering
    \includegraphics[scale=0.12]{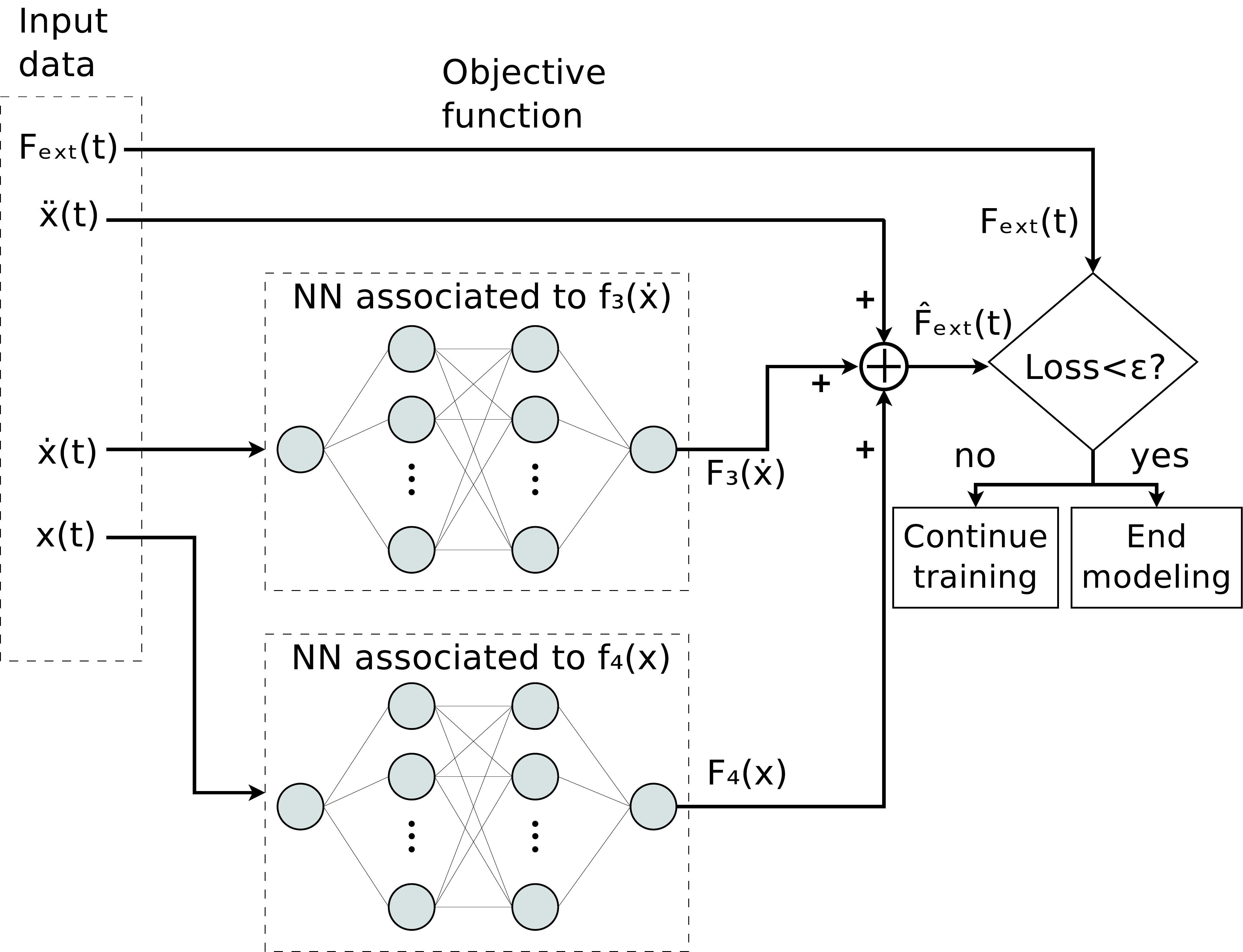}
    \caption{Schematic diagram for NN-CC implementation of the velocity-dependent friction model.}
    \label{fig:NN_structure:vel}
\end{figure}

It is important to mention a comment about the uniqueness of the NN-based decomposition, first considering the position-dependent friction model. 
Suppose that after two independent trainings, we obtain two pairs of networks $[F_1(x(t);\theta_1) \,,\, F_2(x(t);\theta_2)]$ and $[\tilde{F}_1(x(t);\tilde\theta_1) \,,\, \tilde{F}_2(x(t);\tilde\theta_2)]$, both of which successfully represent the observed dynamics. Then, it is satisfied 
\begin{equation}
 \ddot{x}+F_1(x;\theta_1)\, \dot{x}+F_2(x;\theta_2)= \\
    \ddot{x}+\tilde{F}_1(x;\tilde\theta_1)\, \dot{x}+\tilde{F}_2(x;\tilde\theta_2) \; .
\end{equation}
Rearranging, we obtain
\begin{equation}
[F_1(x;\theta_1)-\tilde{F}_1(x;\tilde\theta_1)]\, \dot{x}+[F_2(x;\theta_2)-\tilde{F}_2(x;\tilde\theta_2)]= 0 \;.
\end{equation}
In phase-space terms, $x$ and $\dot{x}$ act as independent coordinates (real trajectories typically explore a two-dimensional manifold, so $\dot{x}$ is not a single-valued function of $x$). Consequently, the only solution on the sampled domain is $F_1(x;\theta_1)=\tilde{F}_1(x;\tilde\theta_1)$ and $F_2(x;\theta_2)=\tilde{F}_2(x;\tilde\theta_2)$. 
Importantly, this uniqueness implies that the CCs are identical even though their intrinsic parameterizations may differ (i.e., $\theta_1\neq\tilde\theta_1$ and $\theta_2\neq\tilde\theta_2$), emphasizing that the form of the CCs, rather than their parameterizations, are the relevant quantities. 

Concerning the case of the velocity-dependent friction model [Eq.~\ref{eq:model:veloc}], where $f_3$ depends only on $\dot{x}$ and $f_4$ only on $x$, the two functions are mathematically independent except for an additive constant. Specifically, any constant term can be arbitrarily reassigned between $f_3$ and $f_4$, making the decomposition non-unique. To resolve this ambiguity: (i) In the Poly-CC and SINDY-CC approaches, we simply omitted the constant term in $f_4$; (ii) In the NN-CC method, 
we imposed the constraint $f_4(x)=0$ during training by adding a term $\lambda_c F_4(x)$ to the loss function, with $\lambda_c\ll1$ (we chose $\lambda_c=0.01$). 
This constraint ensures that any constant contribution is adsorbed in $F_3$, leading to a unique decomposition.

This functional independence also has a clear physical interpretation. Suppose that the system includes a damping element (e.g., a viscous friction) and a restoring element (e.g., a spring). If we modify only the damping (e.g., by increasing friction), while the spring remains unchanged, then only the function representing the damping should vary, while the spring-related function remains unaffected. This implies that the identification procedure is capable of recovering each element consistently and independently.

\end{itemize}

\section{Results}\label{sec:results}
Two representative examples are presented in this section: the van der Pol oscillator (Sec.~\ref{sec:van_der_pol}), and a stick-slip system (Sec.~\ref{sec:stickslip}). 
All numerical simulations in this study, including the theoretical simulations of the governing ODEs and the simulations for the three approaches implemented, were conducted using the LSODA (Livermore solver for ordinary differential equations with automatic method switching for stiff and non-stiff problems) integration method\cite{Hindmarsh1983,Petzold1983}. LSODA was selected due to its demonstrated superior numerical stability and efficiency compared to standard explicit methods such as the Runge-Kutta-Fehlberg (RK45) algorithm, especially when dealing with systems characterized by stiffness, chaotic behavior, or certain types of discontinuities where the solution changes rapidly\cite{William2007numericalrecipes}.  
For each simulation, we verified that the values of $\left[\ddot{x}+f_1(x)\,\dot{x}+f_2(x)-F_{ext}(t)\right]^2$ (for model 1) and $\left[\ddot{x}+f_3(\dot{x})+f_4(x)-F_{ext}(t)\right]^2$ (for model 2) were lower than $10^{-16}$ at each time instant, thus verifying that the systems were correctly integrated. 
For all simulations in these examples, we simulate forward up to $t_{max}$=40 seconds and  then interpolate the solution to obtain 500 values uniformly spaced ($N_{steps}$=500).  
Accordingly, each dataset used for system identification consists of 500 uniformly sampled data points for each one of the functions $x(t)$, $\dot{x}(t)$, $\ddot{x}(t)$, and $F_{ext}(t)$.

For the NN-CC approach, we employed a NN architecture with two hidden layers, each comprising 100 neurons and rectified linear unit (ReLU) activation functions (see Appendix C for details). 
This architecture was chosen to balance the flexibility and computational efficiency of the model. This setup is consistent with common architectures in PINNs\cite{Raissi2019}, which often use shallow networks with hundreds of neurons for approximating physical systems.  
According to the universal approximation theorems, a feedforward NN architecture with more than one hidden layer and a sufficient number of neurons can approximate any continuous\cite{Cybenko1989,Pinkus1999} or piecewise continuous\cite{Hornik1989} function with arbitrary precision, given enough neurons and hidden layers (see also \cite{Goodfellow2016} for recent approaches).  We trained the NNs using the Adam optimizer, without regularization or input/output normalization. In our tests, this straightforward setup converged reliably in both continuous and discontinuous cases. Introducing L1 or L2 regularization degraded performance in systems with discontinuities, as they tend to smooth out sharp transitions. Furthermore, normalizing inputs/outputs to [–1, 1] yielded no appreciable benefit. Although alternative strategies and architectures can change metrics slightly, they do not alter the main findings reported here. Training was performed for up to 20.000 epochs or until the loss error reached $10^{-7}$, whichever was satisfied first. All models reached final loss errors below $10^{-4}$, indicating good convergence. 
We have applied a linear extrapolation scheme for evaluating the NN-CC models beyond the training data range, as detailed in Appendix~A.


For the Poly-CC approach, we used polynomial expansions of the CCs up to order 10. Higher orders led to instabilities in cases with discontinuous functions, while lower orders failed to capture the sharp transitions accurately.

Similarly, in the SINDy-CC approach, the candidate library included polynomial terms up to order 10. 
Using higher-order terms resulted in instability during forward simulations for discontinuous systems. We also tested other candidate functions, such as trigonometric terms, but they provided no significant improvement. Therefore, we report results only for the polynomial basis.




\subsection{Example 1: van der Pol}\label{sec:van_der_pol}
\begin{figure}[htpb]
    \centering
    \includegraphics[width=6cm]{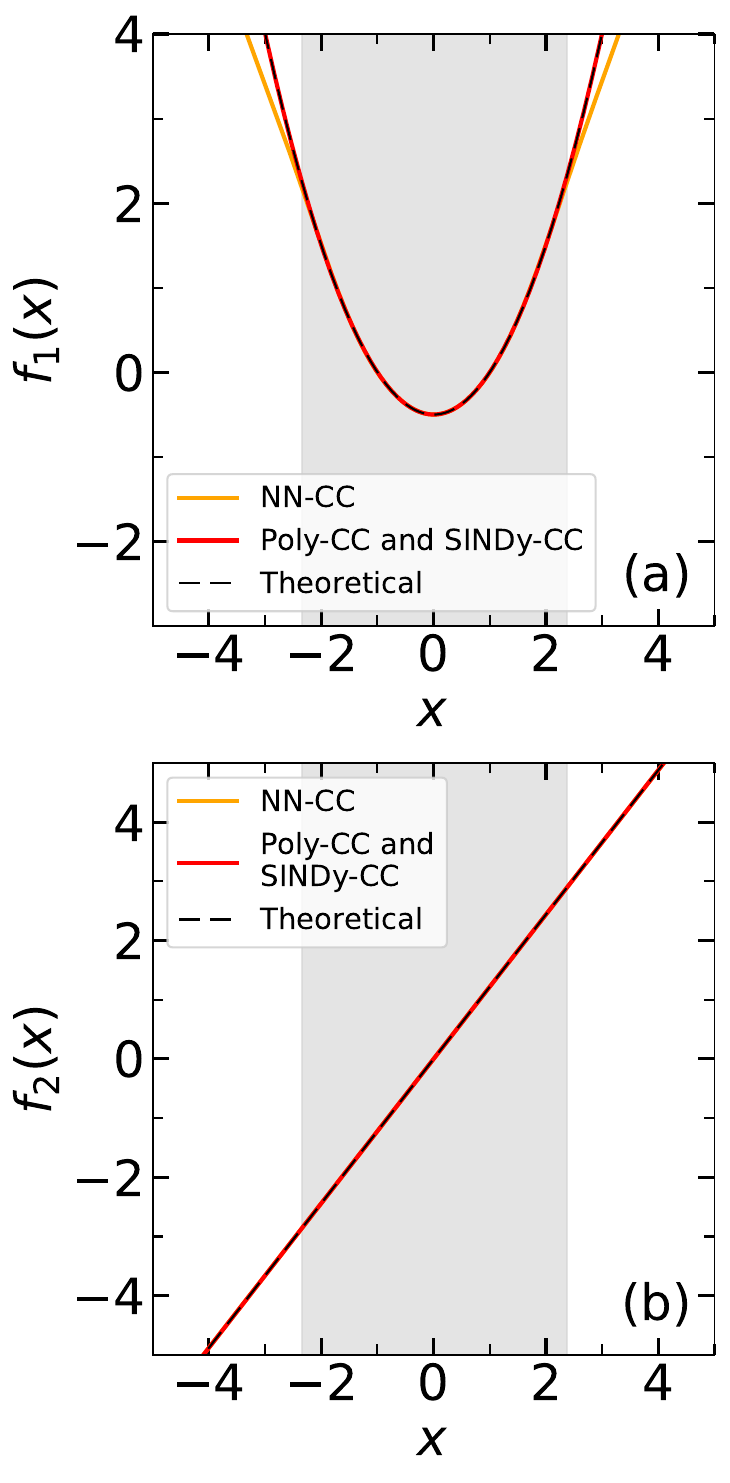}
    \caption{Obtained characteristic curves for the van der Pol example using NN-CC, Poly-CC and SINDy-CC methods: (a) $f_1(x)$, and (b) $f_2(x)$. The shaded gray regions indicate the range of training data used for identification. Obtained NN-CC model  (orange solid lines),  Poly-CC and SINDy-CC model are visually identical (represented both with red solid lines), and theoretical CCs (black dashed lines). Models were evaluated on an extended region beyond the gray shaded region. Parameters:
    $A=0.834$, $k=1.22$, $\mu=0.501$, $\Omega=1.512$, $x_0=-0.353$ and $v_0=-0.408$.}
    \label{fig:van_der_pol:cc}
\end{figure}

The van der Pol system can be described by Eq.~\ref{eq:model:pos}, where the CCs are defined as
\begin{equation}
    f_1(x)=\mu\;(x^2-1) \;\; \text{and} \;\; f_2(x)=k \, x \;,
    \label{eq:cc:van_der_pol}
\end{equation}
and the external forcing is given by $F_{ext}(t)=A\, \cos(\,\Omega\; t\,)$. 
We first consider an example in which the theoretical ODE is numerically integrated using the parameters $A=0.834$, $k=1.22$, $\mu=0.501$, $\Omega=1.512$, with initial conditions $x_0=-0.353$ and $v_0=-0.408$. This simulation gives us a dataset (DS) composed of $\{x(t), \dot{x}(t), \ddot{x}(t), F_{ext}(t)\}$, which is used for system identification using the three approaches: NN-CC, Poly-CC, and SINDy-CC.
The obtained CCs after the identification process are shown in Fig.~\ref{fig:van_der_pol:cc}, along with the gray shaded region that represents the range of $x$ values for the training dataset. Predictions beyond this region correspond to extrapolations of the models.  
All three methods show similar values for the CCs within the gray shaded region. However, notable discrepancies appear outside this area because NN-CC employs linear extrapolation, as detailed in Appendix~A. We chose this specific approach for its simplicity; however, more complex schemes may be used for practical applications if necessary.



\begin{figure*}[htpb]
    \centering
    \includegraphics[width=15cm]{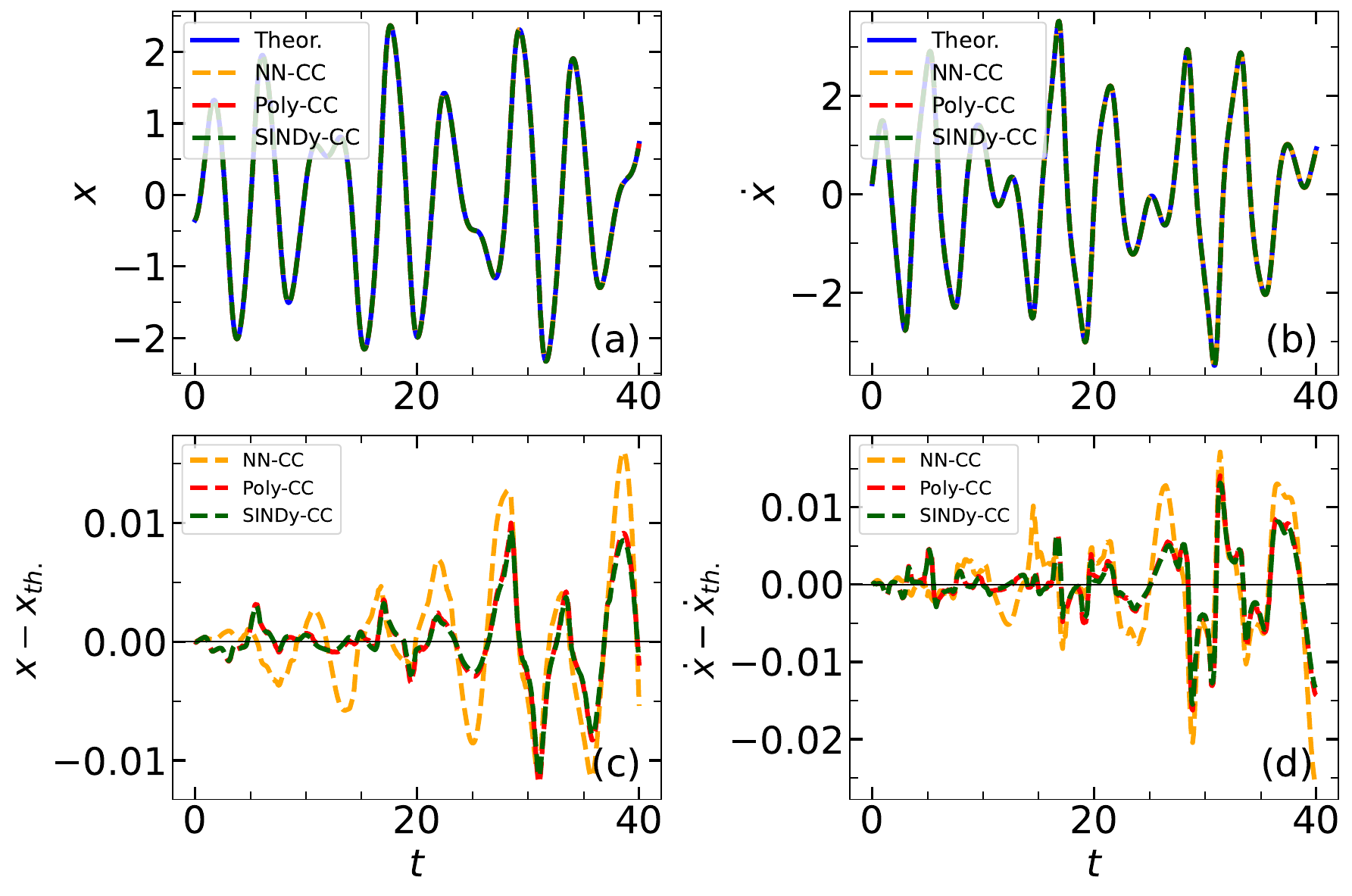} 
    \caption{Validation example of different models obtained from a dataset with $A=0.834$, $k=1.22$, $\mu=0.501$, $\Omega=1.512$, $x_0=-0.353$ and $v_0=-0.408$, and here validated for other initial conditions ($x_0=-0.385$ and $v_0=0.213$) 
    and driven force ($A=1.384$, $\Omega=1.578$): (a) and (b) show $x(t)$ and $\dot{x}(t)$ obtained from theoretical, NN-CC, Poly-CC and SINDy-CC forward integrations; (c) and (d) show the error with respect to the theoretical simulation for $x(t)$ and $\dot{x}(t)$, respectively. }
    \label{fig:van_der_pol:val}
\end{figure*}

As a validation example, the ODE was integrated forward using a new initial condition and forcing parameters: $x_0=0.458$, $v_0=0.033$, $A=1.384$, and $\Omega=1.578$, while keeping the model parameters fixed at $\mu=0.501$ and $k=1.22$. The theoretical dynamical variable and its derivative are shown in Fig.~\ref{fig:van_der_pol:val}, together with the simulations predicted by the previously identified models.
Panels (a) and (b) show the good agreement between the theoretical and predicted values of $x(t)$ and $\dot{x}(t)$, respectively. To provide a more detailed assessment, panels (c) and (d) display the corresponding differences with respect to the theoretical solution: $x(t) - x_{th}(t)$ and $\dot{x}(t) - \dot{x}_{th}(t)$. 
To quantify the accuracy of each method, we computed the RMSE of $x(t)$ with respect to the theoretical solution as
\begin{equation}
   \text{RMSE}[method]=\left(\frac{1}{N_{\text{data}}}\sum_{i=0}^{N_{\text{data}}-1} [x(i\, h)-x_{th}(i\, h)]^2\right)^{1/2}
\end{equation}
where $h=t_{max}/N_{\text{data}}$ denotes the time step size and $x(i\, h)$ corresponds to the dynamical variable of the corresponding \textit{method}. The resulting RMSE values for this validation example were: RMSE[NN-CC]=$4\cdot10^{-3}$, RMSE[SINDy-CC]=$3.1\cdot10^{-3}$, and RMSE[Poly-CC]=$3.3\cdot10^{-3}$.  


\begin{figure}[htpb]
    \centering
    \includegraphics[scale=0.35] {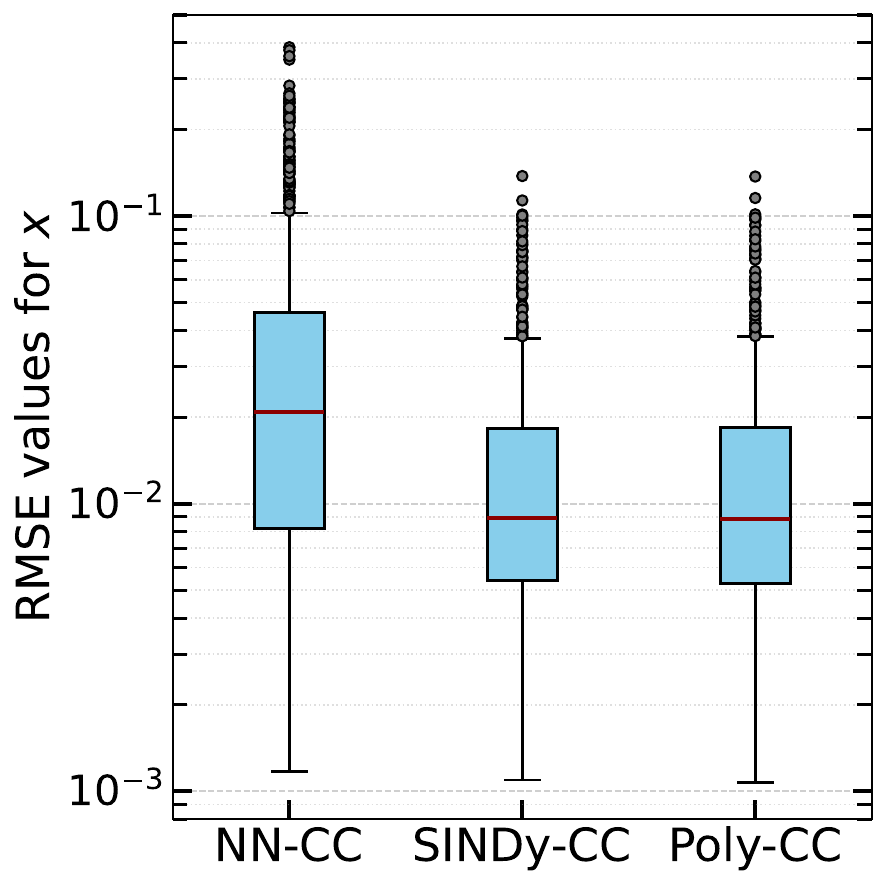}
    \caption{RMSE values for the validation of the van der Pol oscillator, based on 900 forward simulations, reported for the three CC-based approaches.}    \label{fig:van_der_pol:rmse}
\end{figure}

To assess the generalization capabilities of the identified models, we validated the models under different initial conditions and external forcing. 
Hence, to ensure a comprehensive exploration of the phase space, we generated 30 training datasets by randomly sampling the parameters $\mu$, $k$, $A$, $\Omega$ and initial conditions $x_0$ and $v_0$, from uniform distributions within the following intervals: $\mu\in[0.5,10]$, $k\in[0.5,1.5]$, $A\in[0,2]$, $\Omega\in[0,5]$, $x_0\in[-0.5,0.5]$ and $v_0\in[-0.5,0.5]$. 
Each of these training datasets was used to obtain the system model using the three identification methods. 
For validation, we generated 30 additional datasets for each training dataset by maintaining fixed the parameters $\mu$ and $k$ to the values used in the corresponding training DS, while randomly sampling new values for $A$, $\Omega$, $x_0$ and $v_0$ in the same intervals as above. 
This procedure resulted in 30 training DSs and 30 associated validation DSs per each training, yielding a total of $30\times30=900$ RMSE values per modeling method. The statistical distributions of these RMSE values are summarized in the box-plots shown in Fig.~\ref{fig:van_der_pol:rmse}. 
The model parameters obtained using SINDy-CC and Poly-CC were consistent with the underlying governing equation, with discrepancies smaller than $10^{-16}$ for the $\mu$ and $k$ parameters, demonstrating excellent reconstruction accuracy.


Among the three approaches, NN-CC method exhibited a slightly higher RMSE value, which can be attributed to two primary factors: (i) the high number of degrees of freedom in the NNs, which may introduce minor inaccuracies in the CCs that can accumulate and result in appreciable deviations; and (ii) extrapolation errors that occur when the system is simulated under conditions in which $x(t)$ exceeds the range of values encountered during the model identification stage (i.e., beyond the shaded gray region in Fig.~\ref{fig:stick_slip:cc}). A more detailed discussion of these extrapolation issues is provided in Appendix~A.

Additionally, Appendix~B explores a hypothetical scenario in which the underlying dependency (position- or velocity-dependent friction model) is unknown a priori. In such cases, applying both models and validating their predictions under distinct driving forces or initial conditions provides a method to select the correct model. 
This cross-validation approach is typically sufficient to determine which model (or whether neither of the considered models) accurately captures the system dynamics.

\subsection{Example 2: dry friction (stick-slip)}\label{sec:stickslip}

\begin{figure}[htpb]
    \centering
    \includegraphics[width=6cm]{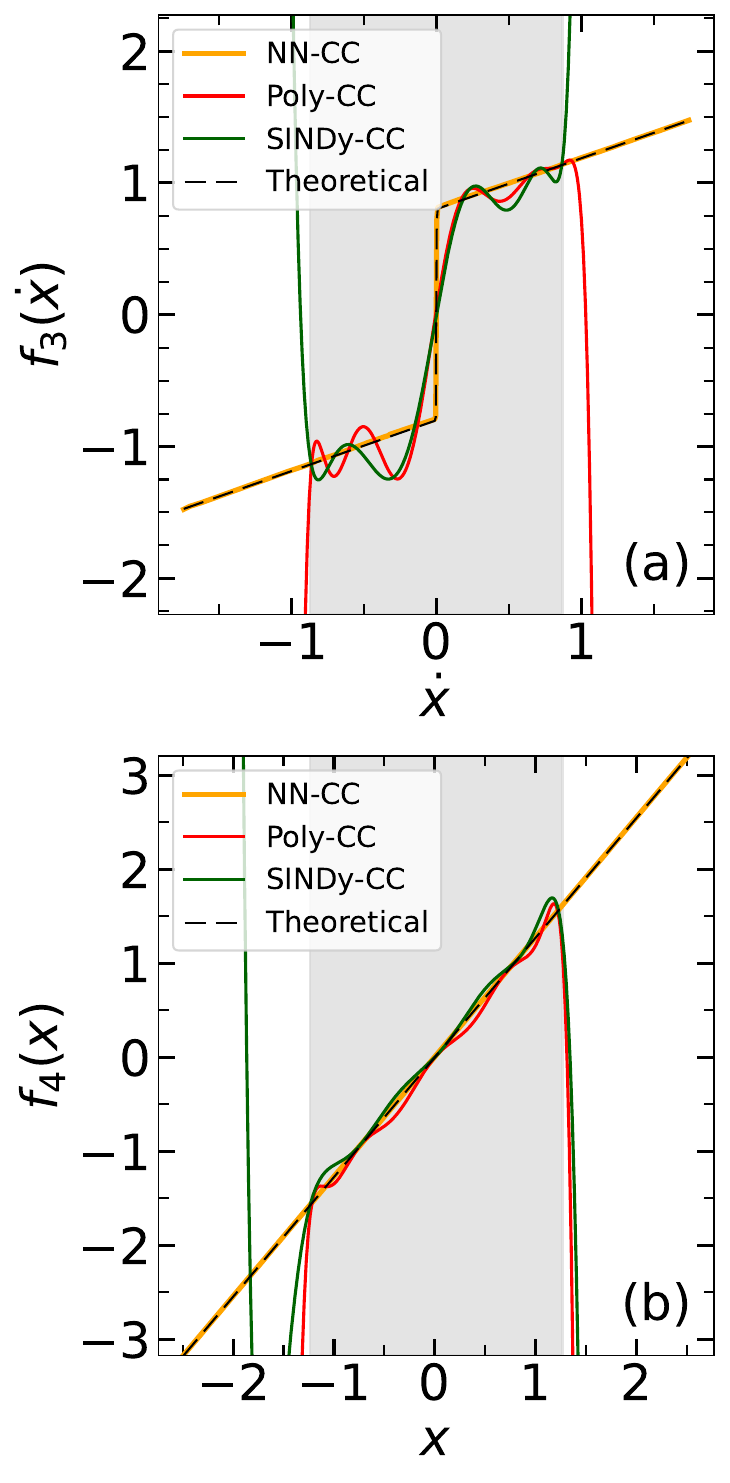}
    \caption{Obtained characteristic curves for the van der Pol example using NN-CC, Poly-CC and SINDy-CC methods: (a) $f_1(x)$, and (b) $f_2(x)$. The shaded gray regions indicate the range of training data used for identification. Obtained NN-CC model (orange solid lines),  Poly-CC model (red solid lines), SINDy-CC model (green solid lines), and theoretical CCs (black dashed lines). Models were evaluated on an extended region beyond the gray shaded region. Parameters: $x_0=-0.076$ , $v_0=0.146$, $\mu\,.\,N=0.801$, $A=2$, $k=1.274$, $\Omega=0.363$, $c=0.386$.}
    \label{fig:stick_slip:cc}
\end{figure}

The stick-slip system with Coulomb friction can be described by Eq.~\ref{eq:model:veloc}, where the CCs are defined as
\begin{equation}
    f_3(x)= c\;\dot{x}+ \mu \;N \; \text{sign}(\dot{x})\;\; \text{and} \;\; f_4(x)=k \, x \;,\label{eq:cc:stick_slip}
\end{equation}
and the external forcing is given by $F_{ext}(t)=A\, \cos(\,\Omega\; t\,)$. 
We first consider an example in which the theoretical ODE is numerically integrated using the parameters $A=2$, $k=1.274$, $c=0.386$, $\Omega=0.363$, $\mu\cdot N=0.801$, $x_0$=-0.076, and $v_0$=0.146. 
This simulation yields to a training dataset (DS) composed of $\{x(t), \dot{x}(t), \ddot{x}(t), F_{ext}(t)\}$. 
The obtained CCs after the identification process are shown in Fig.~\ref{fig:stick_slip:cc}, along with the gray shaded region that represents the range of $x$ values for the training dataset. Predictions beyond this region correspond to model extrapolations.


Poly-CC and SINDy-CC present oscillations at the discontinuity, and extrapolation issues are observed beyond the gray shaded region. As discussed in Refs.~\cite{Gonzalez2023,Gonzalez2024}, these effects arise from their intrinsic polynomial structure. Specifically, oscillations arise from Runge phenomena and Gibbs oscillations, while divergences in the extrapolated regions beyond the edges result from the inherently divergent nature of high-degree polynomials.

Remarkably, the NN-CC did not exhibit oscillatory artifacts, and the extrapolation challenge was effectively mitigated through the application of the linear extrapolation criterion, as detailed in Appendix~A.

For Poly-CC and SINDy-CC, this divergent behavior resulted in instabilities during forward simulations. To mitigate these issues,  we selected initial conditions and external forcings within a narrower range. This ensured that the training datasets encompassed wider range values for $x$ and $\dot{x}$ compared to the range used for validation.

\begin{figure*}[htpb]
    \centering
    \includegraphics[width=15cm]{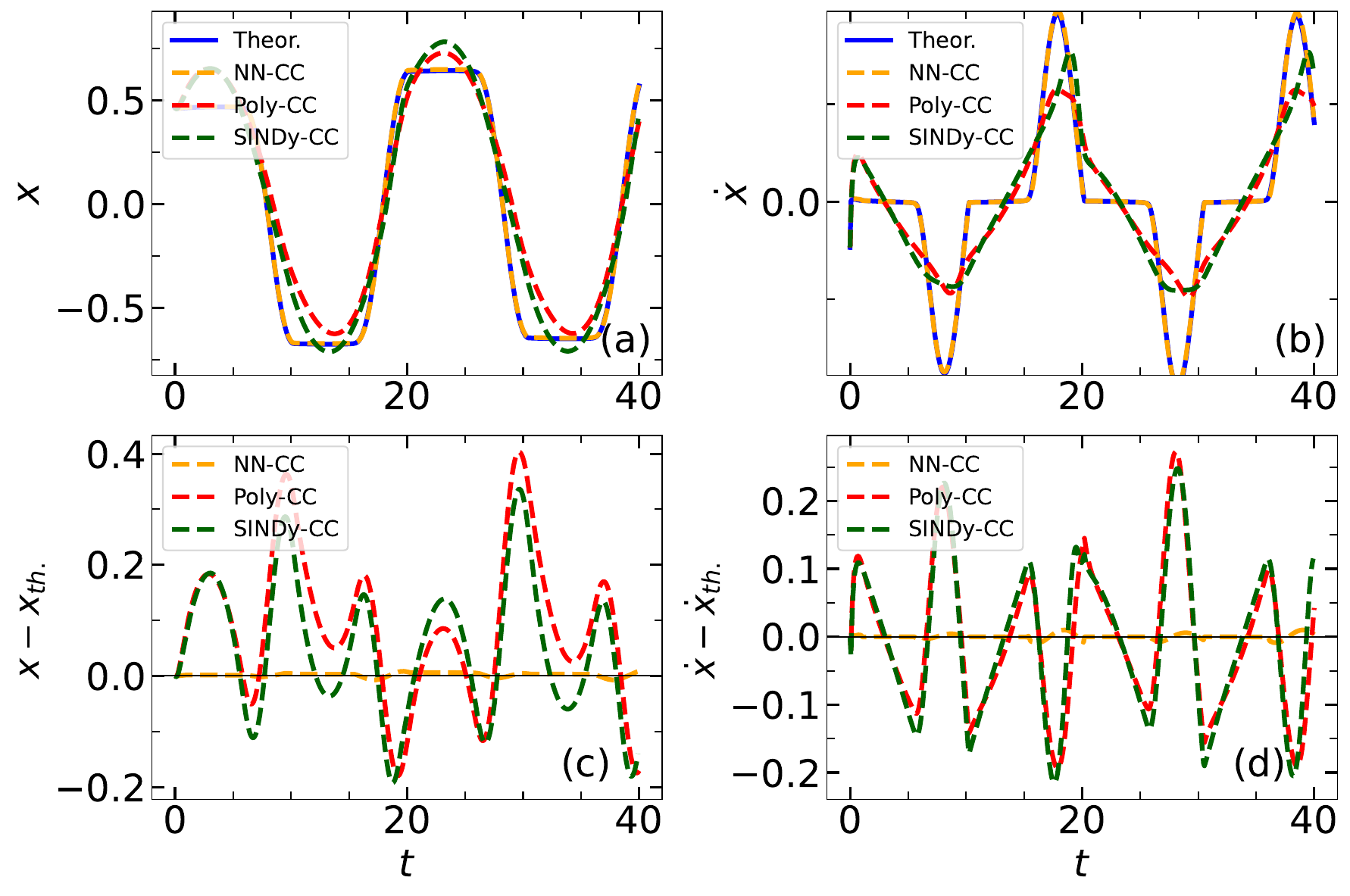} 
    \caption{Validation example for stick-slip system obtained from a dataset with
    Parameters: $x_0=-0.076$ , $v_0=0.146$, $\mu\,.\,N=0.801$, $A=2$, $k=1.274$, $\Omega=0.363$, $c=0.386$, 
    and here validated for other initial conditions ($x_0=0.458$ and $v_0=0.033$) and external force ($A=1.384$, $\Omega=0.578$): (a) and (b) show $x(t)$ and $\dot{x}(t)$ obtained from theoretical, NN-CC, Poly-CC and SINDy-CC forward integrations; (c) and (d) show the difference with respect to the theoretical simulation for $x(t)$ and $\dot{x}(t)$, respectively. }
    \label{fig:stick-slip:val}
\end{figure*}

To illustrate a validation example, Fig.~\ref{fig:stick-slip:val} shows the simulation results using parameters $A=1.396$, $\Omega=0.306$, and initial conditions $x_0$=0.4640 and $v_0$=-0.1170. 
Both SINDy-CC and Poly-CC approaches exhibit significant deviations from the true system dynamics, primarily because of their inadequate representation of the CCs near the discontinuity. 
The resulting RMSE values are: RMSE(NN-CC)=$5\cdot10^{-3}$, RMSE(SINDy-CC)=$1.3\cdot10^{-1}$, and RMSE(Poly-CC)=$1.6\cdot10^{-1}$. 

To sample a broad range of configuration and parameter spaces, we generated 30 training datasets by randomly selecting the parameters $\mu\cdot N$, $k$, $c$, $\Omega$ and $x_0$ and $v_0$ from uniform distributions within the following intervals: $\mu\cdot N\in[0.5,1]$, $k\in[0.5,1.5]$, $c\in[0.1,0.5]$, , $\Omega\in[0,5]$, $x_0\in[-0.5,0.5]$ and $v_0\in[-0.5,0.5]$, while keeping fixed $A=2.0$. 
Subsequently, for the validation process, we generated 30 datasets for each training dataset, by maintaining fixed the model parameters $\mu\cdot N$, $k$, $c$ at the values of used in the corresponding training DS, while the parameters $\Omega$, $x_0$, $v_0$ were randomly selected from the same uniform intervals as mentioned above, and $A$ was varied  in a narrowed range within $[1,1.5]$.  
By selecting this range of variation for the validation datasets, we ensured that the simulated $x(t)$ trajectories remained within the gray-shaded region corresponding to each training dataset, thereby avoiding the instabilities observed for the Poly-CC and SINDy-CC simulations.

\begin{figure}[htpb]
    \centering
    \includegraphics[scale=0.35]{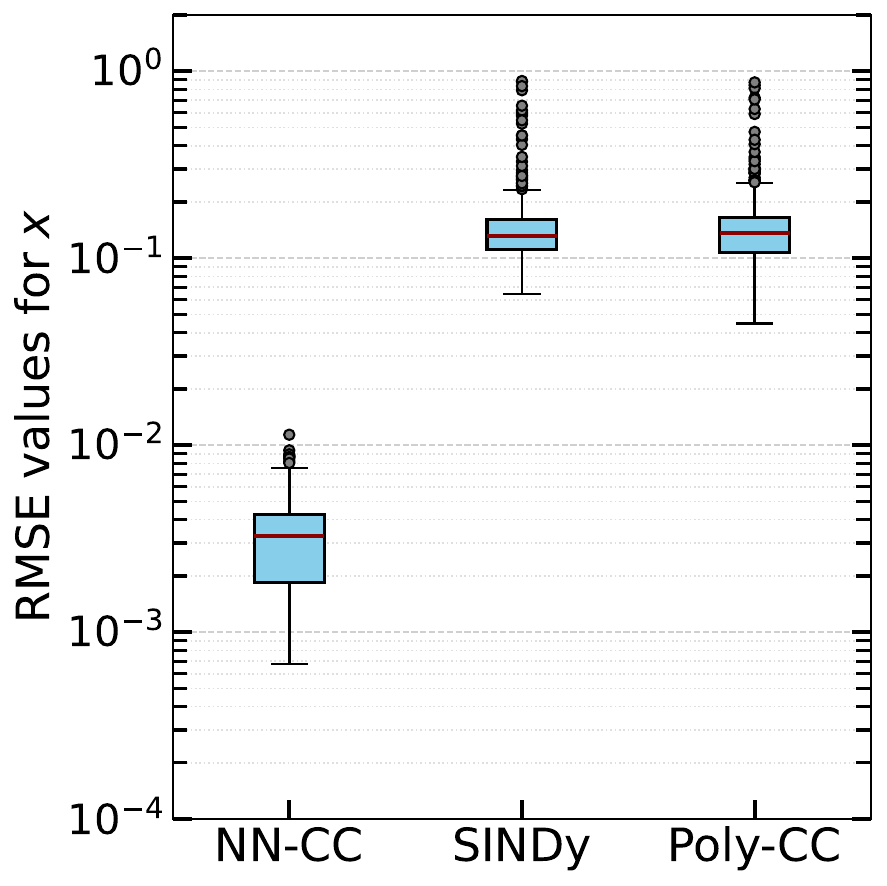}
    \caption{RMSE values for the validation of the stick-slip system, based on 900 forward simulations, reported for the three CC-based approaches. }
    \label{fig:stick-slip:rmse}
\end{figure}

In total, this procedure yielded 30 independently generated training DSs and 30 validations per each training, resulting in $30\times30=900$ RMSE values per modeling method. The box-plots for these RMSE values are presented in Fig.~\ref{fig:stick-slip:rmse}. 
SINDy-CC and Poly-CC exhibit similar performance, while the NN-CC method performs significantly better.
Quantitatively, the NN-CC method achieved a mean RMSE of $\approx$ 50 times lower compared to the results of SINDy-CC and Poly-CC, with RMSE(NN-CC)=$3\cdot10^{-3}$, RMSE(SINDy-CC)=$1.6\cdot10^{-1}$, and RMSE(Poly-CC)=$1.5\cdot10^{-1}$.

\section{Discussion}\label{sec:discussion}

System identification methods typically involve trade-offs between flexibility,
interpretability, and reliance on prior knowledge of the system (such as functional forms, symmetries, and physical constraints). 
Black-box models provide high flexibility and predictive performance, but often lack interpretability. 
Physics-informed neural networks (PINNs) incorporate known physical laws into the learning process (e.g., by including the EDOs into the loss function), yet the system dynamics remain embedded within a monolithic NN, making it difficult to extract functional relationships or interpret model parameters. 
Sparse identification techniques (e.g., SINDy) offer interpretable models but require a predefined library of candidate functions, which can limit flexibility if the true system dynamics cannot be captured by the chosen basis. Symbolic regression methods can discover functional relationships by only specifying a set of mathematical operators but they are typically computationally intensive and do not yield a unique model, often returning multiple candidate expressions with similar performance.


In this work, we introduce the characteristic curve (CC)-based framework for identifying two families of second-order nonlinear dynamical systems. By isolating the fundamental components of the dynamics (namely, friction and restoring forces) as explicit functions $f_i$ ($i=\{1,\,\ldots\,,4\}$), the method enables a structured and physically interpretable modeling approach. 
Hence, a key strength of the CC-based framework lies in its interpretability, especially when contrasted with black-box models such as deep neural networks, recurrent architectures, neural ordinary differential equations (neural ODEs), and physics informed neural networks (PINNs). While these modern techniques have shown remarkable performance in data fitting and forecasting tasks, they often lack transparency and make it difficult to extract mechanistic insights about the system. In contrast, the CC-based approach offers a clear decomposition of the dynamics into identifiable building blocks: position-dependent friction, velocity-dependent friction, and elastic forces. Each of these functions can be visualized, analyzed, and compared directly to known physical laws or empirical models.

This interpretability is not only philosophically appealing, instead, it has practical implications. For instance:
\begin{itemize}
\item[(i)] It allows domain experts to validate or reject specific physical hypotheses (e.g., whether damping is linear, quadratic, or piecewise).
\item[(ii)] It provides a foundation for model reduction, by enabling the practitioner to approximate CCs using simplified or parametric forms.
\item[(iii)] It makes the models transparent and auditable, which is essential in safety-critical or regulated applications (e.g., mechanical systems, biomedical devices).
\item[(iv)] It allows symbolic regression or physical law discovery techniques to be easily integrated or guided.
\end{itemize}
Although (ii) and (iv) were not explored in detail in this work, they are natural directions to be addressed in future studies.

Furthermore, the CC framework provides an explicit representation of the governing equation in terms of observable quantities. This transparency is expected to reduce the risk of overfitting, facilitate better generalization, and assist practitioners in defining an adequate model extrapolation beyond the training domain by analyzing the partially identified CCs. Although all of these aspects were not the main focus of the present work, they offer promising directions for future investigation. 
In this study, we implemented a simple linear extrapolation in the NN-CC approach; however, future research may explore more sophisticated extrapolation strategies. 

Another advantage of the CC-based framework lies in the modularity of the approach. Each CC can be treated independently, making it possible to incorporate prior knowledge (e.g., known elastic behavior or friction laws) for some components, while learning others from data. This level of flexibility is difficult to achieve in monolithic black-box architectures where interactions are entangled within a single, often opaque, model. Exploring how to exploit this modularity more systematically could lead to more efficient and interpretable modeling strategies in future work.

From an application perspective, accounting for noise and measurement uncertainty is essential when dealing with real systems.
Future work should evaluate the robustness of the CC-based methods under such conditions, potentially incorporating pre- or post- processing strategies to enhance reliability during CC identification. 
Chaotic systems also deserve dedicated and detailed investigations, which lies beyond the scope of this work. Future research may focus on evaluating the performance of CC-based methods in addressing the specific challenges posed by chaos, such as sensitivity to initial conditions, parameter variability, and long-term prediction accuracy. 


For practitioners who treat the system as a black-box and are primarily concerned with predictive accuracy (regardless of model interpretability), PINNs or other modern deep learning approaches\cite{Chandra2024deep,Chandra2025HistRNN,Pilloneto2025} may indeed offer competitive performance compared to NN-CC. However, a fair and meaningful comparison would require testing on real-world datasets, possibly including noise and uncertainties, and would depend heavily on the specific application. A detailed benchmark of these comparisons lies beyond the scope of the present work but is an interesting future direction.  
PINNs are designed to incorporate the governing equations directly into the loss function, enabling them to learn solutions that satisfy known physical laws. However, the internal function approximator (a neural network) acts as a black-box, and standard PINNs do not aim to recover or expose the functional form of individual components of the dynamics. In contrast, our CC framework is explicitly designed to identify interpretable, component-wise functions that reflect the internal structure of the system. While PINNs are powerful tools for enforcing physical constraints, they are not typically used for model discovery or for interpreting the contribution of individual terms in the dynamics, which is a central motivation of our approach.


Based on the experience gained from the systems studied in this work, we suggest the following guidelines: 
\begin{itemize}
\item Use CC-based methods when model interpretability and grounded ODE discovery are important, as they provide a transparent and practical way to represent the system dynamics by explicitly identifying and visualizing the CCs. 
This approach is particularly useful when the system is known to belong to a family of models, but the exact functional form is uncertain. For instance, in friction systems where it is uncertain whether the friction force follows Coulomb, Dieterich-Ruina, or another form, the NN-CC method can infer the underlying functional form directly from data. 
Additionally, due to the uniqueness of the CC-based representation and the flexibility of NNs, the NN-CC approach can be used to evaluate whether a given dataset is consistent with a proposed family of differential equations. This capability is particularly useful for validating or discarding candidate model families.

\item Explore symbolic regression methods (e.g., PySR\cite{Cranmer2023PySR}) when the mathematical structure is unknown, as they can discover unexpected functional forms from data. These methods operate by specifying a set of mathematical operators (e.g., $+$, $-$, $\times$, $\div$, $\sin$, $\log$, $\mathrm{abs}$), and typically return a Pareto front of candidate models that balance complexity and accuracy. While offering flexibility, symbolic regression is computationally intensive and does not produce a unique solution, often yielding multiple models with comparable performance. This makes it particularly suitable for offline identification tasks, where identification time is not critical.

\item Consider sparse regression methods (e.g., SINDy\cite{brunton2016}) when the goal is to discover compact, interpretable equations from data, particularly when the underlying dynamics are expected to consist of linear combinations of simple functions, such as polynomials, trigonometric terms, or products of state variables. 
\item Consider using physics encoded sparse identification methods (e.g., PhI-SINDy\cite{Lathourakis2024}) when part of the governing equations is known in advance and should be enforced during model discovery. Phi-SINDy is particularly useful when combining measured data with known physical laws, guiding sparse regression to improve model accuracy and physical consistency while retaining interpretability.

\item Consider using PINNs or other black-box deep learning methods \cite{Raissi2019, Karniadakis2021, Sun2023}) when predictive accuracy is the main goal and the practitioner is comfortable treating the system as black box. These approaches are well suited for scenarios where partial knowledge of the governing equations is available and can be encoded into the training loss, such as through differential equations, constraints, or symmetries. They are particularly useful in complex systems where the explicit functional form is unknown or difficult to model and interpretability is not a priority. Although PINNs can enforce physical consistency and perform well even with limited or noisy data, they do not provide explicit, interpretable representations of the individual components of the dynamics, as the learned behavior remains embedded in the parameters of a NN. Thus, unlike CC-based methods, PINNs are not intrinsically designed for model discovery or for analyzing the contribution of specific terms to the system equations.
\end{itemize}

In this work, we considered three complementary approaches within the CC-based framework: SINDy-CC, Poly-CC, and NN-CC, each offering a different trade-off between interpretability, flexibility, and data-fitting capacity. (i) SINDy-CC provides enhanced interpretability by not only recovering the CCs but also identifying their analytical forms as a sparse combination of predefined basis functions, making it particularly suitable when the underlying dynamics is expected to be governed by linear combination of basis functions. (ii) Poly-CC relaxes the sparsity constraint while retaining interpretability by fitting each CC with a full polynomial expansion, providing a relatively flexible model with parameters that remain analytically transparent. (iii) NN-CC offers significantly greater flexibility by modeling each CC with NNs, enabling the capture of complex and unknown nonlinearities. 
In future work, we plan to develop hybrid approaches combining NN-CC and SINDy-CC, where each CC is first approximated using NN-CC, then visualized and re-fit using sparse regression to identify a compact, interpretable analytical expression. 
Together, these three approaches form a spectrum of modeling strategies that can be selected (or combined) depending on the available prior knowledge and the complexity of the target system.


Although not addressed in this work, extending the CC-based framework to account for CCs that depend on additional parameters, such as time, temperature, or other environmental factors, as well as expanding the CC-based framework to include higher degrees of freedom, represents an important direction for future research aimed at improving its applicability to real-world systems.


In this paragraph, we report identification times on an Intel Xeon CPU: Poly-CC and SINDy-CC completed in $\sim$20 and $\sim$31 $ms$, respectively, while NN-CC required $\sim$37 $s$ under the same conditions. 
It is important to note that these results correspond to a single-threaded baseline. Ongoing work on parallelizing the NN-CC approach has already achieved promising results, with speedups greater than fivefold on multi-core CPU and GPU architectures. Further optimizations are underway and will be reported in future work.



\section{Conclusions}\label{sec:conclusions}


The primary goal of this work is to explore the feasibility and advantages of the characteristic curve (CC) formalism for constructing structured and interpretable models of two families of second-order dynamical systems (position- and velocity-dependent friction). 
We focus on demonstrating how the CC framework enables modular, transparent, efficient, and physically meaningful representations of nonlinear dynamics. 
A key contribution is the demonstration of the uniqueness of these CC-based model representations, meaning that for a given system and dataset, the identified characteristic curves correspond to a single, well-defined mathematical description of the underlying dynamics.
This CC decomposition enables each component of the model to be directly associated with interpretable physical components (e.g., damping or restoring forces), providing a clear advantage in applications where model transparency and insight into system behavior are essential. 
Moreover, the visualization of the CCs provides insight into the system behavior at the boundaries of the training data, helping to define reasonable extrapolation strategies for forward simulations beyond the training region.
This robustness makes the framework suitable for a broad range of nonlinear dynamical systems. Nonetheless, future research should carefully consider the effects of noise, measurement uncertainty, and parameter variability to assess the applicability in practical environments.


We proposed and compared three complementary approaches within the CC-based framework, each offering different trade-off between parameter interpretability and model flexibility: (i) SINDy-CC represents each CCs as a sparse combination of functions from a predefined library, making it well-suited for systems governed by a combination of simple basis functions; (ii) Poly-CC employs full polynomial expansions for each CC, preserving interpretability of parameter values while capturing moderately complex nonlinearities; and (iii) NN-CC provides the highest flexibility by modeling CCs with NNs, eliminating the need for predefined basis functions and enabling approximation of complex or unknown nonlinear functions that may be difficult to express analytically.

In summary, we introduced a CC-based framework for identifying two families of second-order nonlinear systems, enabling interpretable and modular representations of complex dynamics through the learned CCs. These CCs enable direct inspection of the underlying dynamics and can be used for forward simulation, model validation, and control design. 
The framework is defined generically for each family, allowing different models to be identified without the need to modify the code for each specific system. Among the methods, NN-CC is particularly well-suited for capturing a wide range of nonlinearities, including discontinuities. 
Future work may focus on automated extrapolation strategies for the NN-CC approach, the combined use of different CC-based techniques, and the extension of this framework to other families of differential equations and systems with multiple degrees of freedom. Additionally, investigating the effects of measurement noise and conducting a more detailed exploration of chaotic dynamics remain desirable directions for future work.

\section*{Appendix A: Extrapolation issues}

When models are used to simulate the system outside the range of the training data, their output depends on how well they extrapolate beyond that range. This section analyzes how each method used in this work (SINDy-CC, Poly-CC, and NN-CC) behaves in such extrapolation settings.

The SINDy method builds a model by selecting terms from a fixed set of candidate functions. If this set includes the correct mathematical terms of the true system such as a polynomial nonlinearity for the van der Pol example, then SINDy can produce a reliable model that continues to behave well outside the training range. 
However, if the correct terms are missing or if the true function is more complex than the chosen basis, such as those including discontinuous CCs, the extrapolation may be inaccurate. In such cases, the model gives wrong predictions because it cannot capture the true behavior beyond the data. 
In this work, we opted to use only polynomial basis functions for the SINDy-CC approach in order to avoid discussing issues related to the non-unique representation of the CCs. A systematic treatment of this limitation may be addressed in future works. 

The Poly-CC method allows us to obtain a unique representation of the CCs using polynomial functions. This approach yields accurate results even in the extrapolated region for systems with simple polynomial nonlinearities, such as the van der Pol oscillator. However, for more complex CCs, polynomials tend to grow quickly outside their fitting range, often leading to very large or unphysical values when extrapolated (See Ref.~\cite{Gonzalez2024}). This is a well-known problem with high-degree polynomials: even a good fit inside the training interval can lead to strong divergence outside it. Therefore, while the model may behave well within the known data, it can become unreliable in extrapolated regions unless special care is taken.

The NN-CC method uses NNs to model the CCs without assuming a fixed functional form. This makes it flexible and suitable for learning complex relationships, including discontinuities. However, NN extrapolation beyond the training data range is generally unreliable. This is because predictions in extrapolated regions can vary significantly depending on the specific training procedure. Consequently, extrapolated values may exhibit unexpected behavior, including flattening or other unpredictable trends.
Importantly, this issue is not necessarily due to overfitting, but rather to the fact that the network was not trained to make accurate predictions outside the range of the training data. 

To address this issue, we have applied a simple extrapolation scheme for the NN-CC approach: we fitted a first-order polynomial (a line) using the predictions of the previously trained NNs near the edges of the data and extended this line beyond the training domain. This keeps the NN outputs continuous and avoids non-physical behaviors or jumps at the edges. However, it is still an approximation and may not follow the true system behavior. For system simulation we employed the NNs inside the training data domain and used the linear extrapolations outside.

\begin{figure}[H]
    \centering
    \includegraphics[width=6cm]{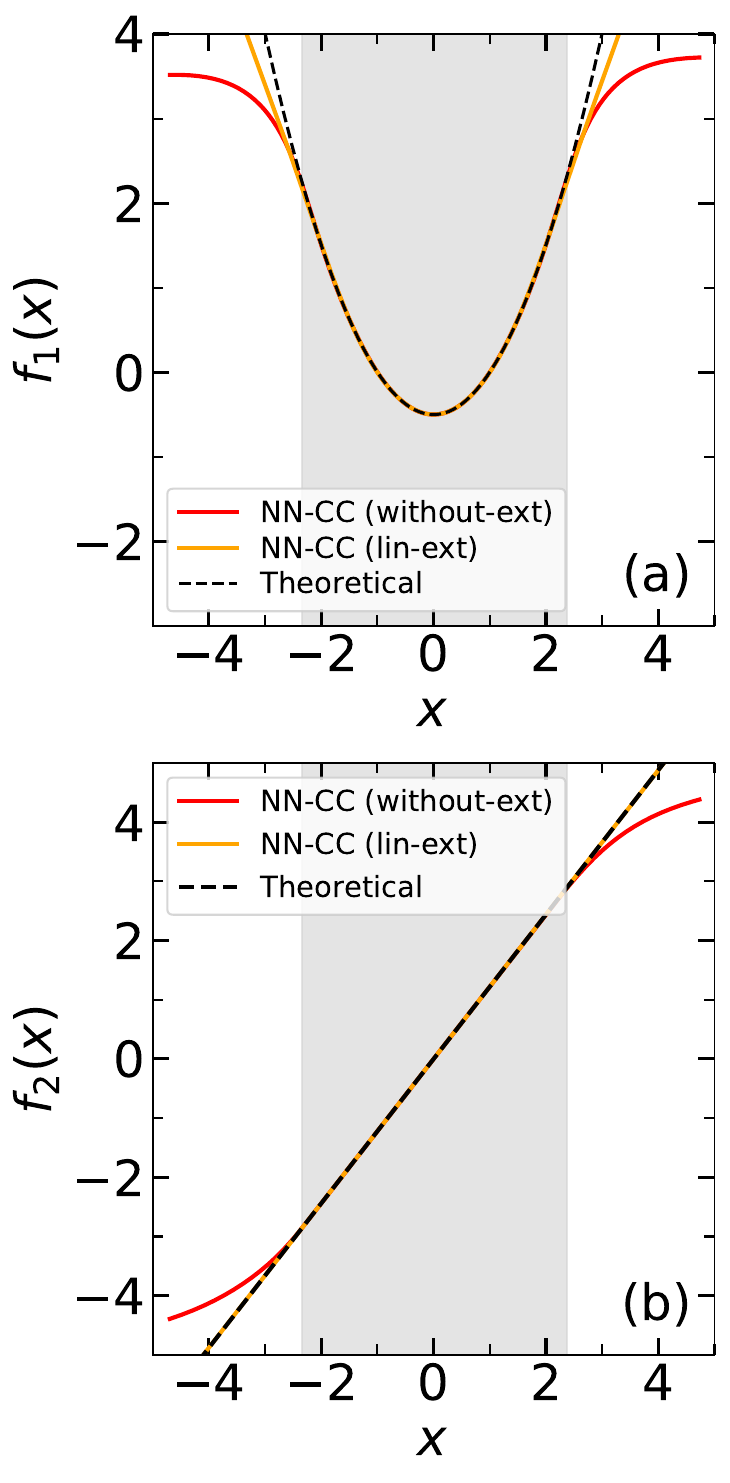}
    \caption{CCs identified for the van der Pol example using the NN-CC method: (a) $f_1(x)$, and (b) $f_2(x)$. The shaded gray regions indicate the range of training data used for identification. The NN-CC model with linear extrapolation (lin-ext) at the edges (orange solid lines), and the direct evaluation of the NNs outside the training region (i.e., without extrapolation) (red solid lines). Theoretical CCs (black dashed lines). Parameters: $x_0=-0.353$ , $v_0=-0.408$, $\mu=0.501$, $A=0.834$, $k=1.22$, $\Omega=1.512$. }
    \label{fig:van_der_pol:cc_appendix}
\end{figure}

\begin{figure}[H]
    \centering
    \includegraphics[width=6cm]{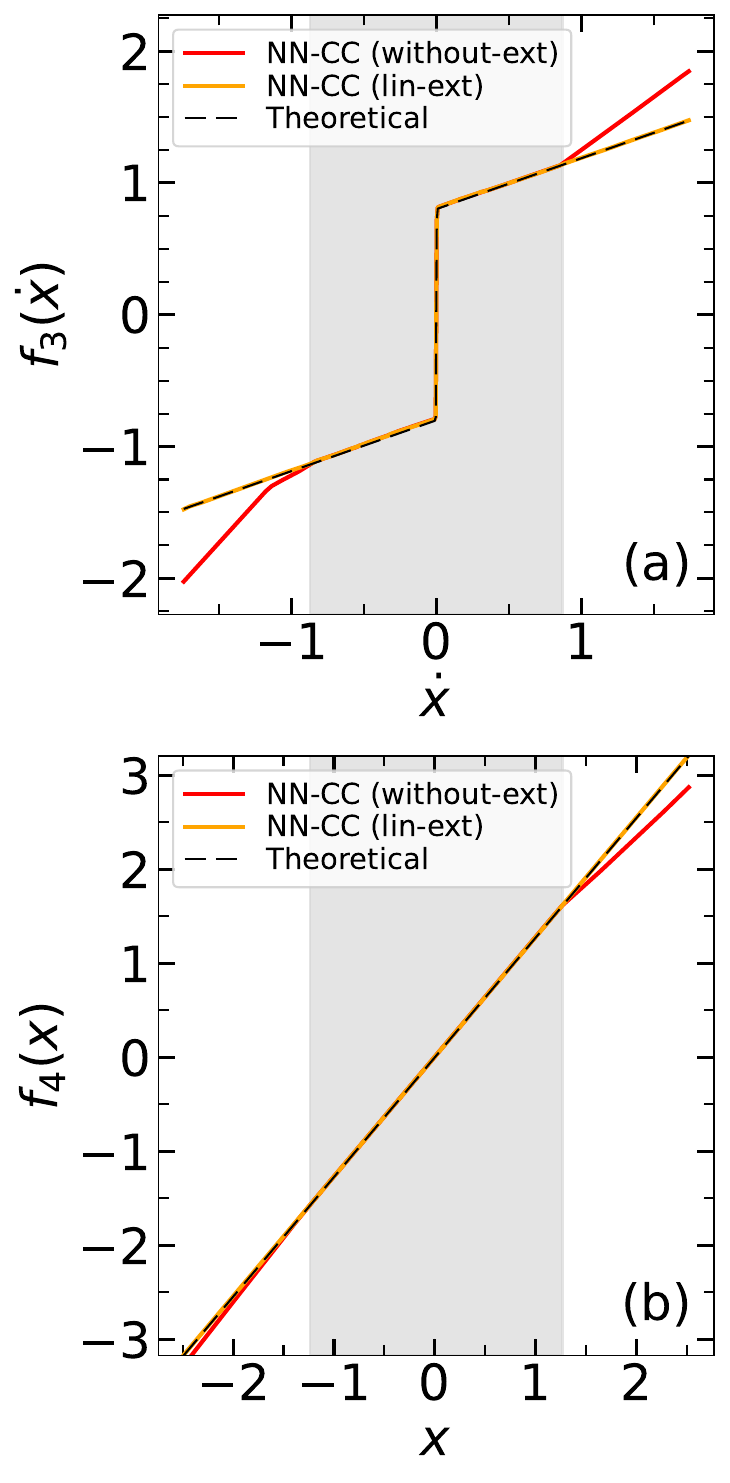}
    \caption{CCs identified for the stick-slip example using the NN-CC method: (a) $f_1(x)$, and (b) $f_2(x)$. The shaded gray regions indicate the range of training data used for identification.  The NN-CC model with linear extrapolation (lin-ext) at the edges (orange solid lines), and the direct evaluation of the NNs outside the training region (i.e., without extrapolation) (red solid lines). Theoretical CCs (black dashed lines). Parameters: $x_0=-0.076$ , $v_0=0.146$, $\mu\,.\,N=0.801$, $A=2$, $k=1.274$, $\Omega=0.363$, $c=0.386$. }
    \label{fig:stick_slip:cc_appendix}
\end{figure}

The effect of evaluating the NNs outside the training regions, as well as the importance of employing an appropriate extrapolation strategy, are shown in Figs.~\ref{fig:van_der_pol:cc_appendix} and \ref{fig:stick_slip:cc_appendix} for the van der Pol and stick-slip systems, respectively. 
These figures demonstrate that the linear extrapolation criterion provides a simple but effective approach for handling extrapolation in both systems.

All three methods have limitations when applied to extrapolation. SINDy-CC performs well if the correct terms are in the basis set. Poly-CC can diverge due to the natural behavior of polynomials outside their fitting range. NN-CC is flexible but can behave unpredictably beyond the training data. In general, it is recommended to use these models within the domain of the data used for training, unless additional constraints or physical knowledge are used to guide their behavior in extrapolated regions.

Although we have employed a simple linear extrapolation for the NN-CC models in this study, alternative strategies may be preferable in specific applications. A general strategy applicable to all SINDy-CC, Poly-CC and NN-CC methods is to analyze the identified CCs and subsequently approximate them using physically motivated functions, which then can be employed for forward simulations. 
For instance, $f_1(x)$ in the van der Pol oscillator exhibits a smooth, low-order nonlinear behavior; therefore, fitting it with a quadratic polynomial is appropriate. This analytic form can then be used for forward simulations, including extrapolation beyond the original data range.  
In the case of stick-slip friction system, the CC suggests a velocity-linear friction with a discontinuity at the origin; accordingly, a Coulomb friction law may provide a natural representation.

Importantly, the NN-CC framework makes no prior assumptions about the basis functions to represent the CCs. This inherent flexibility allows the same methodological approach to be applied across a diverse range of systems without modification. For example, in stick-slip friction systems, 
whether the underlying CCs follow a Coulomb, Dieterich–Ruina, or Stribeck law, the NN-CC approach can be employed directly. Once the CCs have been identified, appropriate functions may be fitted to these curves to facilitate stable and physically meaningful extrapolation. Conversely, the SINDy approach often requires redefining the library of candidate functions for each specific application.

\section*{Appendix B: Identifying the correct model}
The selection of position- or velocity-dependent friction model typically relies on a priori physical insight into the system under study. In the absence of such intuition, however, one cannot readily discern which model (or indeed whether either or both models) faithfully captures the true dynamics. To address this ambiguity, we propose a two‐stage procedure: first, perform system identification using both models with the same training dataset; second, validate each resulting model against an alternative dataset, generated either from different initial conditions or under a distinct external force.

\begin{figure}[H]
    \centering
    \includegraphics[scale=0.4]{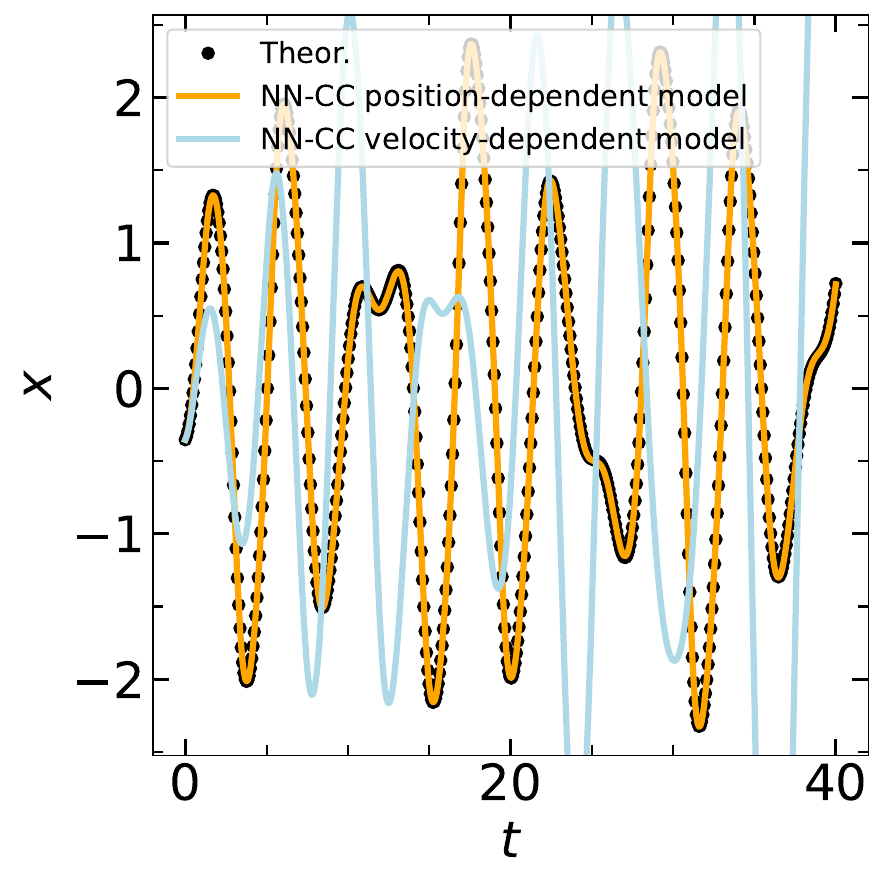} 
    \caption{Validation for the van der Pol system using position- and velocity-dependent friction models. The models were identified using a dataset with $A=0.834$, $k=1.22$, $\mu=0.501$, $\Omega=1.512$, $x_0=-0.353$ and $v_0=-0.408$, and here validated under other initial conditions ($x_0=-0.385$ and $v_0=0.213$) 
    and driven force ($A=1.384$, $\Omega=1.578$).}
    \label{fig:van_der_pol:val_veloc}
\end{figure}

Figure \ref{fig:van_der_pol:val_veloc} presents the validation of the van der Pol oscillator under these two models. Although the velocity-dependent model reproduces the correct initial fragment of the trajectory (i.e., for $t\in [0\,,\,1]$), it rapidly diverges from the theoretical solution, reflecting its inability to encapsulate the governing differential equation. By contrast, the position-dependent model yields a forward simulation that remains in close agreement with the theoretical simulation throughout the validation interval. Consequently, this comparative validation framework provides a criterion for selecting the friction model that most accurately represents the underlying system dynamics. 
If neither model adequately reproduces the validation dataset, this suggests that the underlying model structure is inadequate and must be reconsidered or extended.


\section*{Appendix C: NN architecture}

In this section, we evaluate the performance of different NN architectures and activations functions for the NN-CC method, using a training dataset from the stick-slip system. Table~\ref{tab:number_of_neurons} reports the training error and time as a function of the number of neurons per layer after 20.000 epochs, using two hidden layers and ReLU activation functions. Training errors decrease up to 100 neurons per layer and increases slightly beyond that.  
Therefore, we selected 100 neurons per layer as a good trade-off between model flexibility and training time for this system. 

Table~\ref{tab:hidden_layers} reports the results for different numbers of hidden layers, using 100 neurons per layer and ReLU activation functions. While two layers are sufficient, deeper networks show slightly lower training errors, which may be useful for more complex systems. 

Table~\ref{tab:activation_functions} shows the training error and time for different activation functions, using two hidden layers with 100 neurons per layer. The best performance was achieved with LeakyReLU and ReLU activation functions.

\begin{table}[H]
    \centering
   \caption{{Training errors and times as a function of the number of neurons per layer, using a two hidden-layer architecture. Results were obtained on a single-thread of an Intel Xeon CPU after 20.000 training epochs.}}
    \begin{tabular}{ccc}
    \hline\hline
    Neurons per layer & Loss error & Training time (seg)\\
    \hline 
10   & $9\times 10^{-4}$ &  19 \\
20   & $1.6\times 10^{-4}$ & 23 \\
50   & $4.5\times 10^{-5}$ & 27 \\
100  & $2.5\times 10^{-5}$ & 37 \\
150  & $2.9\times 10^{-5}$ & 55 \\
200  & $3.6\times 10^{-5}$ & 72 \\ 
    \hline\hline 
    \end{tabular}
    \label{tab:number_of_neurons}
\end{table}

\begin{table}[H]
    \centering
   \caption{
   Training error and time as a function of the number of hidden layers, using 100 neurons per layer. Results were obtained on a single-thread of an Intel Xeon CPU after 20,000 training epochs.}
    \begin{tabular}{ccc}
    \hline\hline
    Hidden layers & Loss error & Training time (seg)\\
    \hline 
1   & $6.3\times 10^{-2}$ & 19 \\
2   & $2.5\times 10^{-5}$ & 37 \\
3   & $1.6\times 10^{-5}$ & 55 \\
4  & $9\times 10^{-6}$ & 72 \\
    \hline\hline 
    \end{tabular}
    \label{tab:hidden_layers}
\end{table}

\begin{table}[H]
    \centering
   \caption{Training error and time for different activation functions, using a two-hidden-layer architecture with 100 neurons per layer. Results were obtained on a single-thread of an Intel Xeon CPU after 20.000 training epochs.}
    \begin{tabular}{ccc}
    \hline\hline
    Activation function & Loss error & Training time (seg)\\
    \hline 
LeakyReLU & $2.2\times 10^{-5}$ & 35 \\
ReLU   & $2.5\times 10^{-5}$ & 37 \\
Tanh   & $1.6\times 10^{-4}$ & 37 \\
RRelu   & $9\times 10^{-4}$ & 37 \\
SiLU   & $1.6\times 10^{-3}$ & 45 \\
Sin   & $9.8\times 10^{-3}$ & 39 \\
GeLU   & $1.1\times 10^{-2}$ & 47 \\
SeLU   & $1.1\times 10^{-2}$ & 37 \\
Sigmoid & $1.1\times 10^{-2}$ & 35 \\
SoftPlus & $4.9\times 10^{-2}$ & 51 \\
    \hline\hline 
    \end{tabular}
    \label{tab:activation_functions}
\end{table}

\section*{Appendix D: testing additional position-dependent friction systems with NN-CC}

This appendix demonstrates the application of the NN-CC framework to identify several position-dependent friction models. These models represent a variety of nonlinear systems, illustrating the  flexibility and capability of NN-CC to learn and adapt to different characteristic curves (CCs).

The input dataset for each system studied in this section (defined by the functions $x(t)$, $\dot{x}(t)$, $\ddot{x}(t)$, and $F_{ext}(t)$) was generated via forward integration up to 40 seconds with LSODA method, and then uniformly sampled at 500 data points. Initial conditions were set as $x_0=0.5$, $\dot{x}_0=0.5$.  We then applied the NN-CC approach to obtain the CCs for each system. Importantly, the same code implementation was used across all systems studied in this section, with differences only in the input dataset.


The studied systems were:
\begin{itemize}
\item[(i)] Duffing oscillator:
\begin{equation}
\begin{aligned}
f_1(x)&=\delta\,\\
f_2(x)&=\alpha\,x+\beta\,x^3 \; \,,
\end{aligned}
\end{equation}
with $\delta=0.3$, $\alpha=-1.0$, $\beta=1.0$. Driven force was defined by $F_{ext}(t)=A\,\cos(\omega \, t)$, where $A=0.5$ and $\omega=1.2$. This selection of parameters is known to produce chaotic behavior\cite{Jordan2007}, so this example illustrates that knowing a fragment of the chaotic trajectory is sufficient, in this case, to obtain the characteristic curves (CCs).

\item[(ii)] Impact oscillator:
\begin{equation}
\begin{aligned}
f_1(x)&=c\\
f_2(x)&=k\,x+k\,(x-x_{lim})_+ \; ,
\end{aligned}
\end{equation}
where $z_+=\max(z,0)$, with $c=0.1$, $k=0.5$, and $x_{lim}=0.5$. Driven force was defined by $F_{ext}(t)=A\,\cos(\omega \, t)$, where $A=0.5$ and $\omega=1.2$.

\item[(iii)] FitzHugh–Nagumo model: the standard form of the FitzHugh–Nagumo model is given by:
\begin{align}
\frac{dx}{dt} &= x - \frac{x^3}{3} - w + I_{\text{ext}}(t), \label{eq:fhn_v} \\
\frac{dw}{dt} &= \varepsilon (x + a - b w), \label{eq:fhn_w}
\end{align}
where: $x(t)$ is the fast activator variable (e.g., membrane potential), $w(t)$ is the slow recovery variable, $I_{\text{ext}}$ is an external stimulus current,
$\varepsilon \ll 1$ controls the time scale separation, $a$, $b$ are system parameters. 
We can express Eqs.~\ref{eq:fhn_v} and \ref{eq:fhn_w} as a second-order differential equation in $x(t)$, which is particularly useful for visualizing the CCs:
\begin{equation}
\begin{aligned}
\ddot{x} + ( x^2+\varepsilon\, b- 1 ) \;\dot{x} + \varepsilon \left( a +(1 - b)\,x + \dfrac{b \,x^3}{3}\right) &= - \varepsilon\; b \, I_{ext}(t)\,-\dot{I}_{ext}(t) .
\end{aligned}
\end{equation}
The database was generated with $a=0.7$, $b=0.8$, $\varepsilon=0.08$ and $I_{ext}(t)=A\,\cos(\omega\,t)$, with $A=2$ and $\omega=1.2$.

\end{itemize}

\begin{figure}[H]
    \centering
    \includegraphics[width=0.32\linewidth]{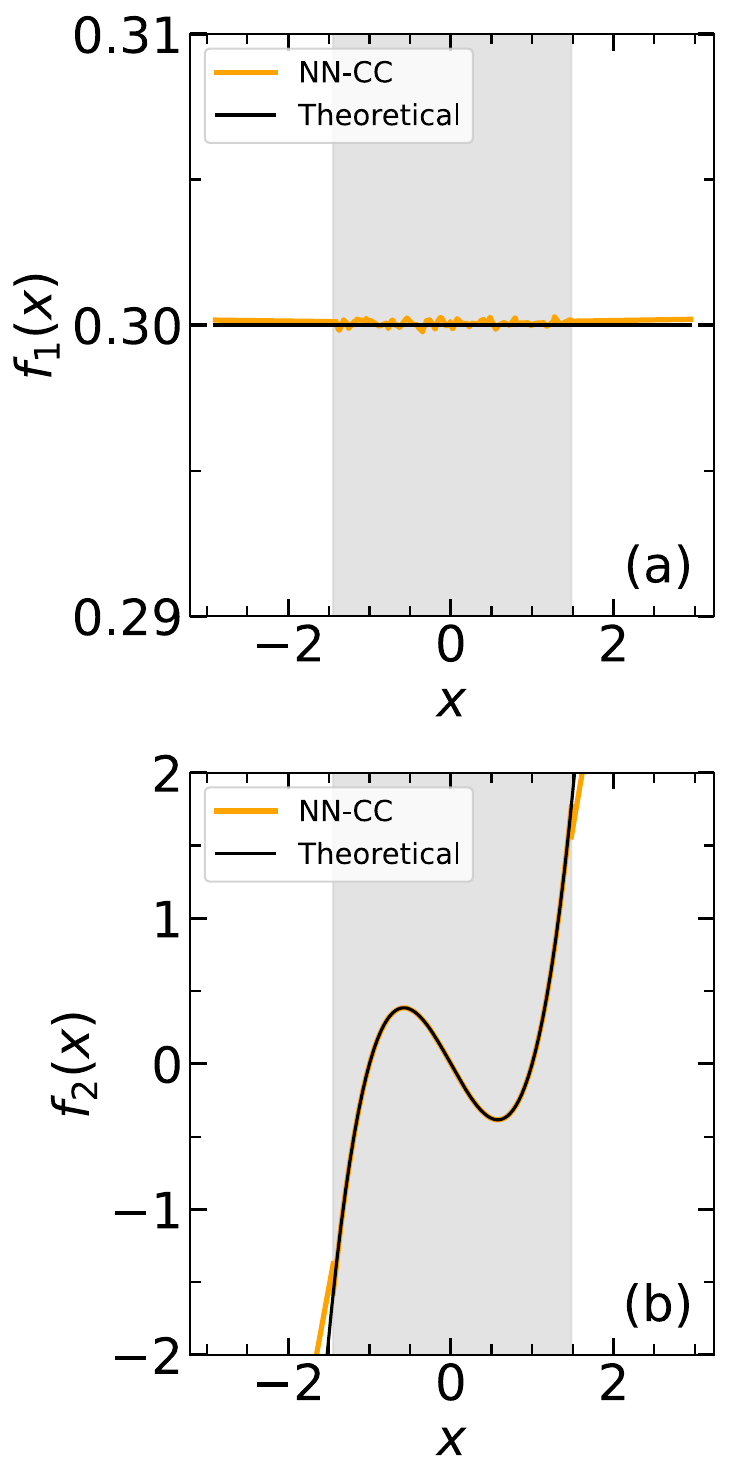}
    \hfill
    \includegraphics[width=0.32\linewidth]{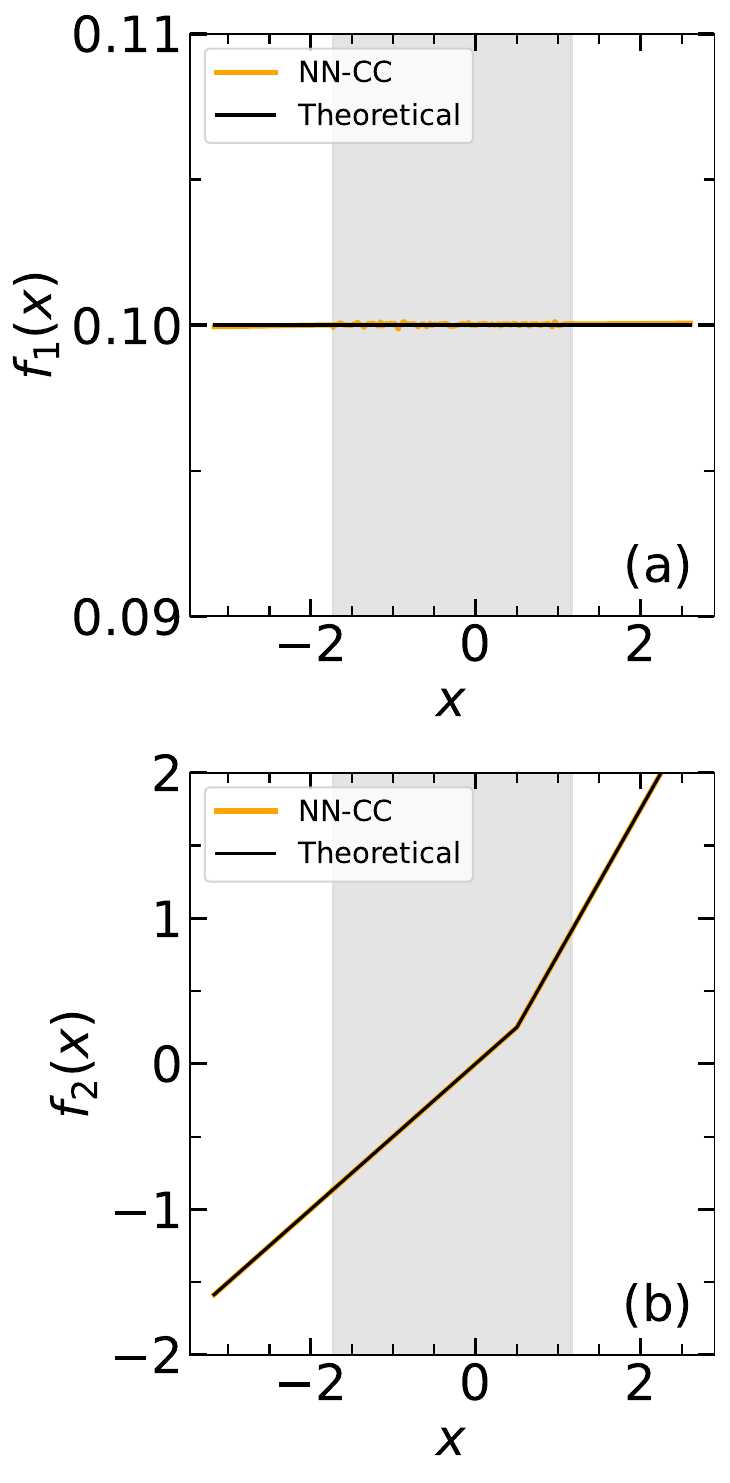}
    \hfill
    \includegraphics[width=0.32\linewidth]{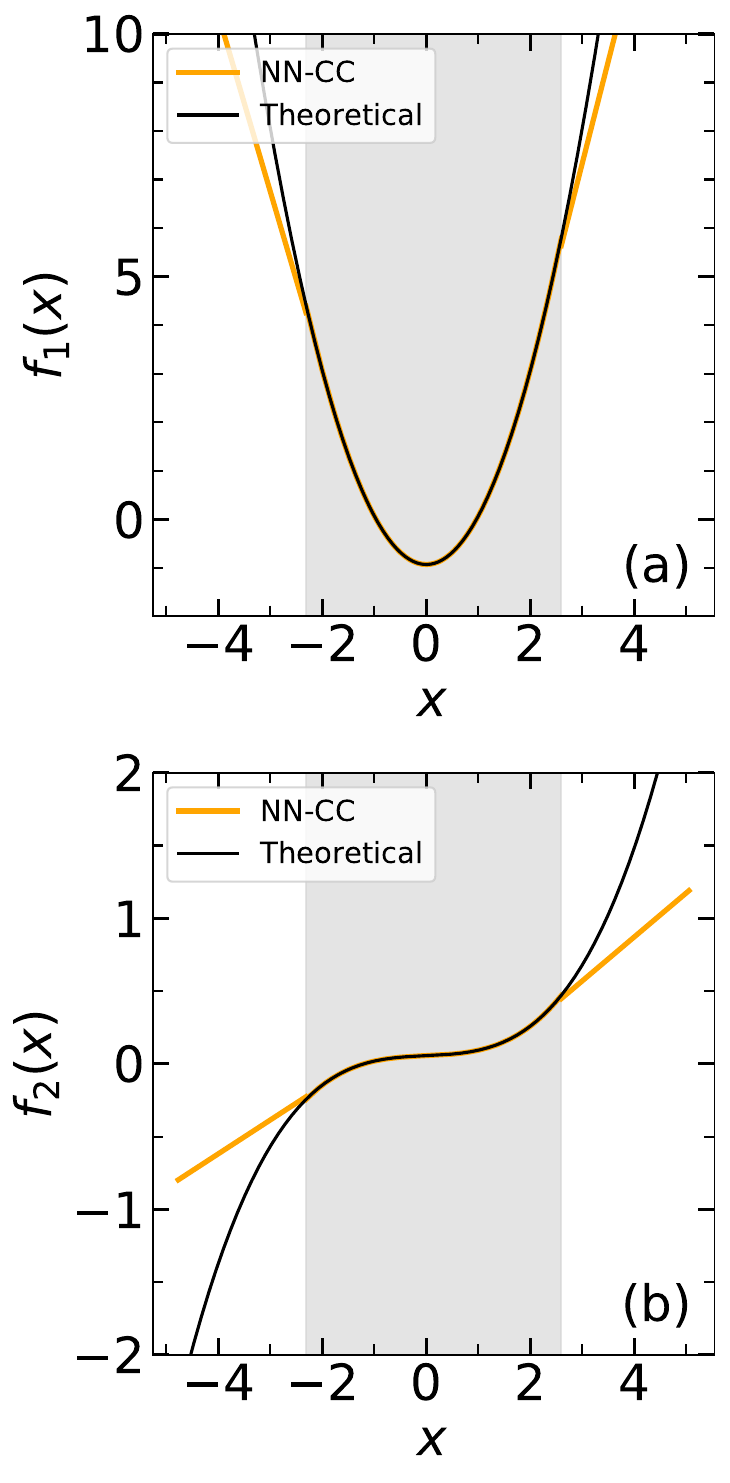}
    \caption{ CCs identified using the NN-CC method for some representative position-dependent friction models: (a) $f_1(x)$ and (b) $f_2(x)$. Results are shown for three representative models: Duffing oscillator (left panels), Impact oscillator (middle panels), FitzHugh-Nagumo (right panels).
    The shaded gray regions indicate the range of theoretical data used for identification. }
    \label{fig:friction-models_1_pos}
\end{figure}

\section*{Appendix E: testing additional velocity-dependent friction systems with NN-CC}

This appendix demonstrates the application of the NN-CC framework to identify several velocity-dependent friction models. These models represent a variety of nonlinear systems, illustrating the  flexibility and capability of NN-CC to learn and adapt to different characteristic curves (CCs).


The input dataset for each system studied in this section (defined by the functions $x(t)$, $\dot{x}(t)$, $\ddot{x}(t)$, and $F_{ext}(t)$) was generated via forward integration up to 40 seconds with LSODA method, and then uniformly sampled at 500 data points. Initial conditions were set as $x_0=0.1$, $\dot{x}_0=0.1$, with a driven force defined by $F_{ext}(t)=A\,\cos(\omega \, t)$, where $A=2$ and $\omega=0.3$.  We then applied the NN-CC approach to obtain the CCs for each system. Importantly, the same code implementation was used across all systems studied in this section, with differences only in the input dataset.


The studied systems were:
\begin{itemize}
    \item[(i)] Dieterich-Ruina:
\begin{equation}
\begin{aligned}
    f_3(\dot{x}) &= c\,\dot{x} + \left[F_f + a \log\left(\dfrac{|\dot{x}| + \varepsilon}{V_f}\right) + b \log\left(d + \dfrac{V_f}{|\dot{x}| + \varepsilon}\right)\right]\;\text{sign}(\dot{x}) \\
    f_4(x)       &= k\,x
\end{aligned}
\end{equation}
with $c=0.1$, $k=1$, $F_f=0.5$, $a=0.07$, $b=0.09$, $d=0.022$, $V_f=3\times 10^{-3}$, and $\varepsilon=1\times 10^{-6}$. These parameters were used recently in Ref.~\cite{Lathourakis2024} within a physics encoded sparse regression approach (PhI-SINDy).

\item[(ii)] Stribeck friction:
\begin{equation}
\begin{aligned}
    f_3(\dot{x})&= c\,\dot{x} + \left[F_f + a \exp\left(-\left(\dfrac{|\dot{x}| }{V_f}\right)^b\ \right)\right]\;\text{sign}(\dot{x})\\
    f_4(x)&=k\,x+\beta \,x^3\;,
\end{aligned}
\end{equation}
with $c=0.1$, $k=1$, $F_f=0.5$, $a=0.07$, and $\beta=0.3$.

\item[(iii)] Coulomb-tanh-power:
\begin{equation}
\begin{aligned}
    f_3(\dot{x})&= c\,\dot{x} + F_f  \;\tanh\left(\alpha \; \dot{x}^\gamma \right)\;\\
    f_4(x)&=k\,x+\beta \,x^3\;,
\end{aligned}
\end{equation}
with  $c=0.1$, $k=1$, $F_f=0.5$, $\alpha=500$, $\beta=0.3$, and $\gamma=3$.

\item[(iv)] Coulomb friction with backlash restoring force:
\begin{equation}
\begin{aligned}
    f_3(\dot{x})&= c\,\dot{x} + F_f  \;\text{sign}\left( \dot{x}\right)\;\\
    f_4(x)&=     \begin{cases}
       k\,x + d  & ;\quad k\,x < -d  \\
       0         & ;\quad |k\,x| \leq d \\
       k\,x - d  & ;\quad k\,x > d  
    \end{cases}\;,
\end{aligned}
\end{equation}
with $c=0.1$, $k=2$, $F_f=0.5$, and $d=0.5$.

\item[(v)] Coulomb friction with piecewise restoring function 1: 
\begin{equation}
\begin{aligned}
    f_3(\dot{x})&= c\,\dot{x} + F_f  \;\text{sign}\left( \dot{x}\right)\;\\
    f_4(x)&=     \begin{cases}
x^2, & x \geq 1 \\
1, & 0.5 \leq x < 1 \\
(x - 1)^3, &  x < 0.5 \\
\end{cases}\;,
\end{aligned}
\end{equation}
with $c=0.1$, and  $F_f=0.5$.

\item[(vi)] Coulomb friction with piecewise restoring function 2: 
\begin{equation}
\begin{aligned}
    f_3(\dot{x})&= c\,\dot{x} + F_f  \;\text{sign}\left( \dot{x}\right)\;\\
    f_4(x)&=     \begin{cases}
0.5 \;x, & x \geq 1.5 \\
x^2, & 1 \leq x < 1.5 \\
1, & 0.5 \leq x < 1 \\
(x - 1)^3, & -0.3 \leq x < 0.5 \\
2x - 3, & x < -0.3
\end{cases}\;,
\end{aligned}
\end{equation}
with $c=0.1$, and  $F_f=0.5$. This example demonstrates the flexibility of the NN-CC approach in accurately modeling a system exhibiting three jump discontinuities.
\end{itemize}

\begin{figure}[H]
    \centering
    \includegraphics[width=0.32\linewidth]{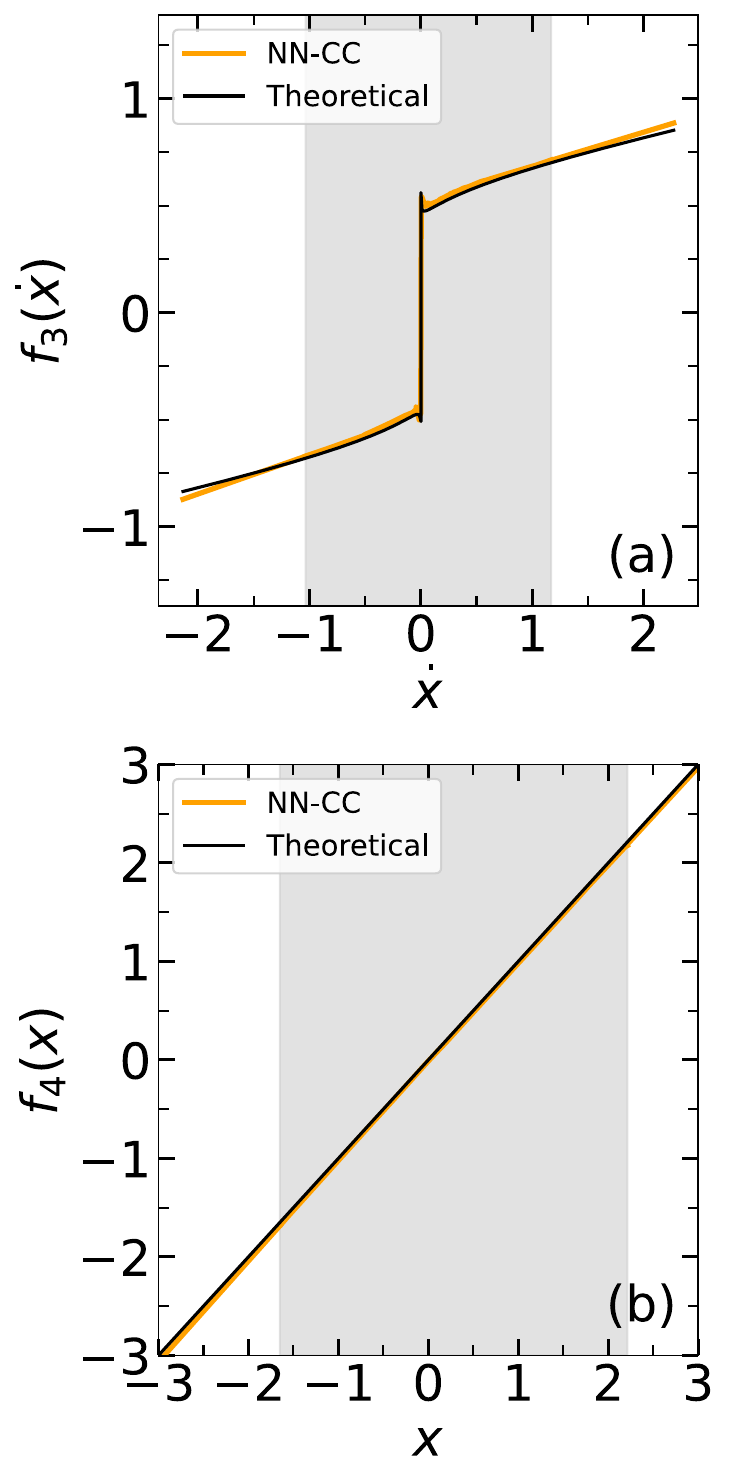}
    \hfill
    \includegraphics[width=0.32\linewidth]{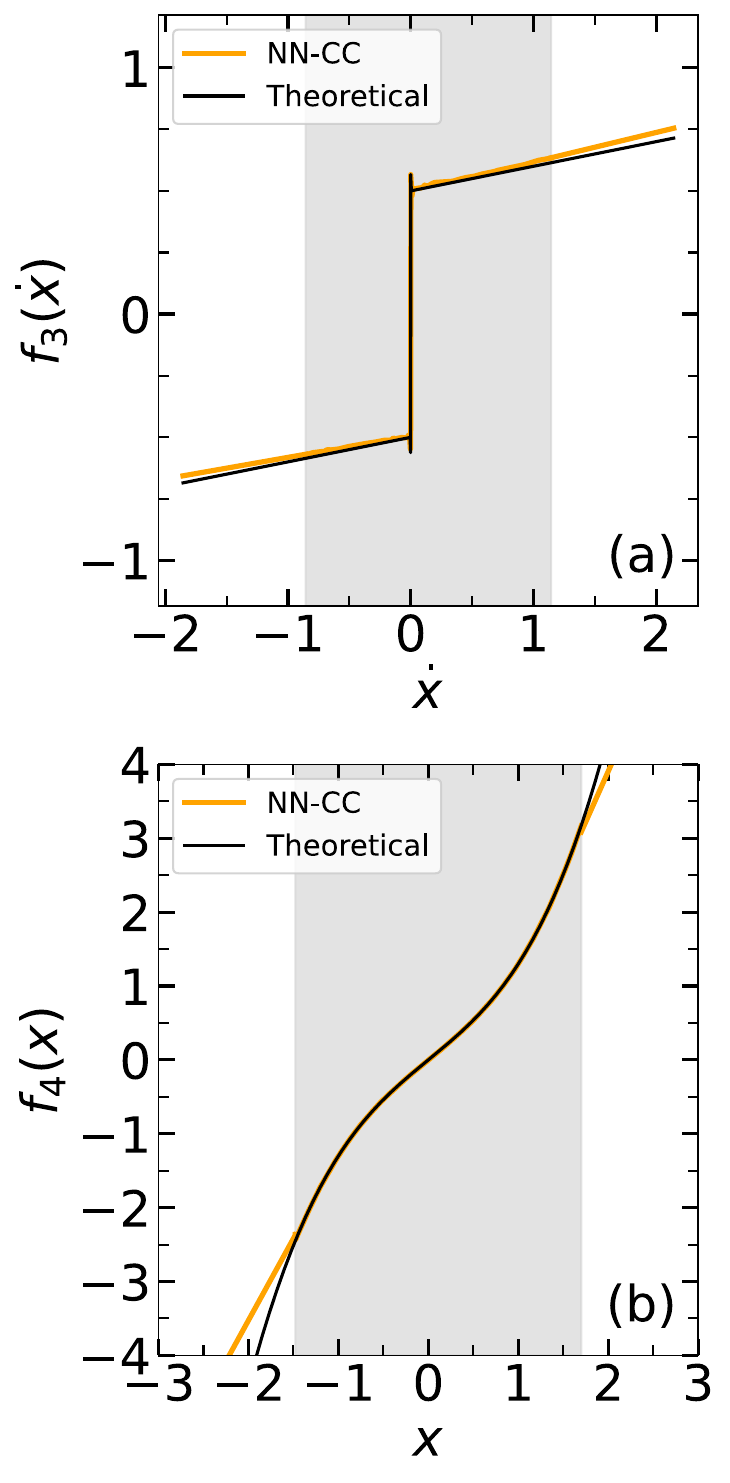}
    \hfill
    \includegraphics[width=0.32\linewidth]{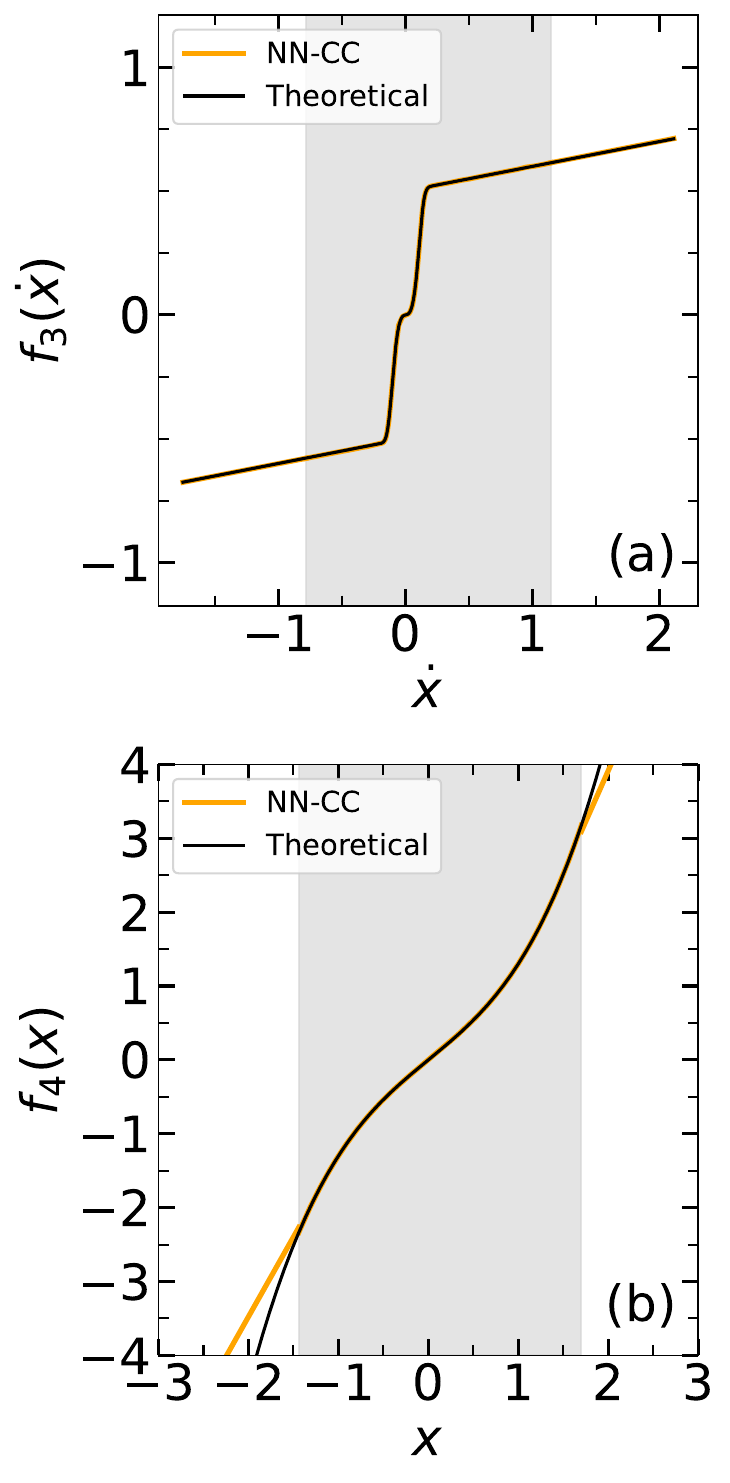}
    \caption{ CCs identified using the NN-CC method for some representative velocity-dependent friction models: (a) $f_3(\dot{x})$ and (b) $f_4(x)$. Results are shown for three representative models: Dieterich-Ruina (left panels), Stribeck (middle panels), Coulomb-tanh-power (right panels).
    The shaded gray regions indicate the range of theoretical data used for identification. }
    \label{fig:friction-models_1}
\end{figure}

\begin{figure}[H]
    \centering
    \includegraphics[width=0.32\linewidth]{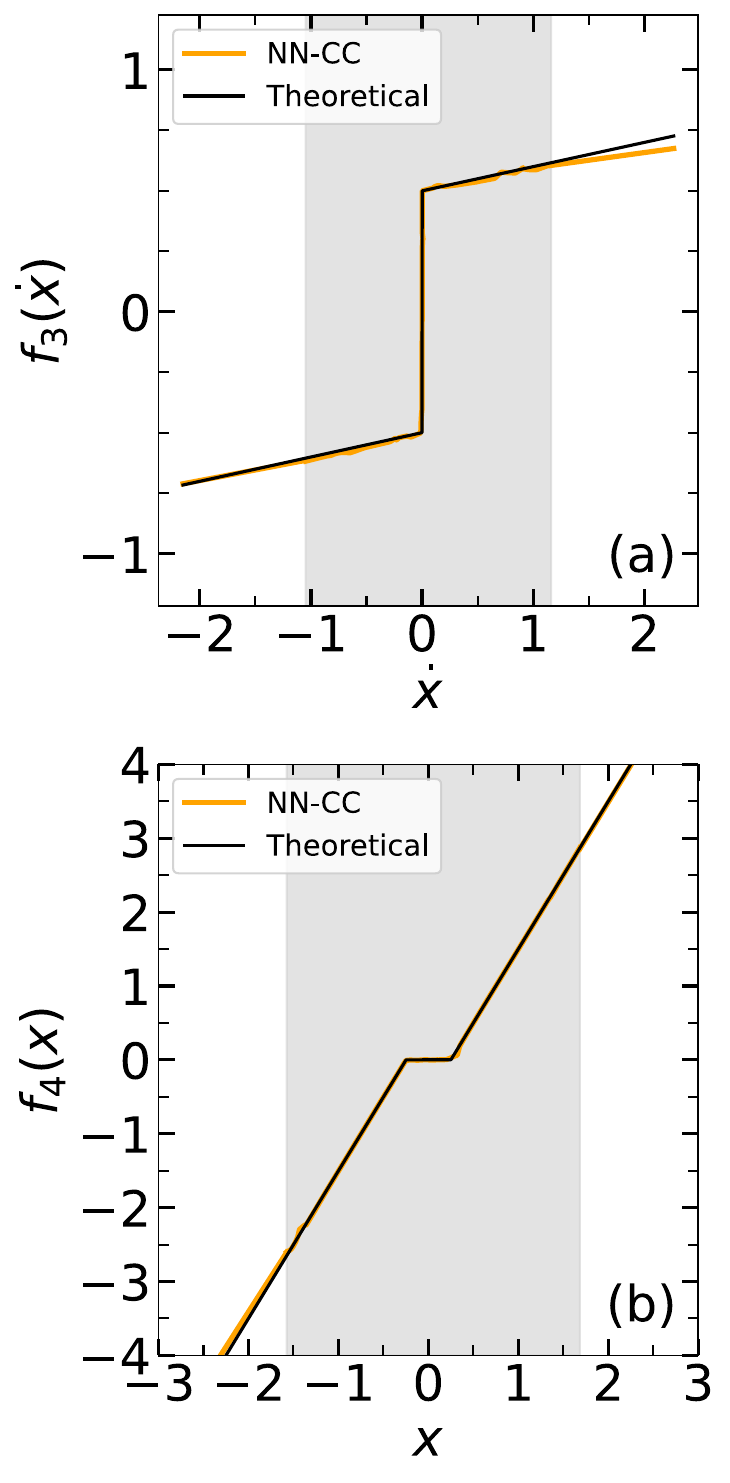}
    \hfill
    \includegraphics[width=0.32\linewidth]{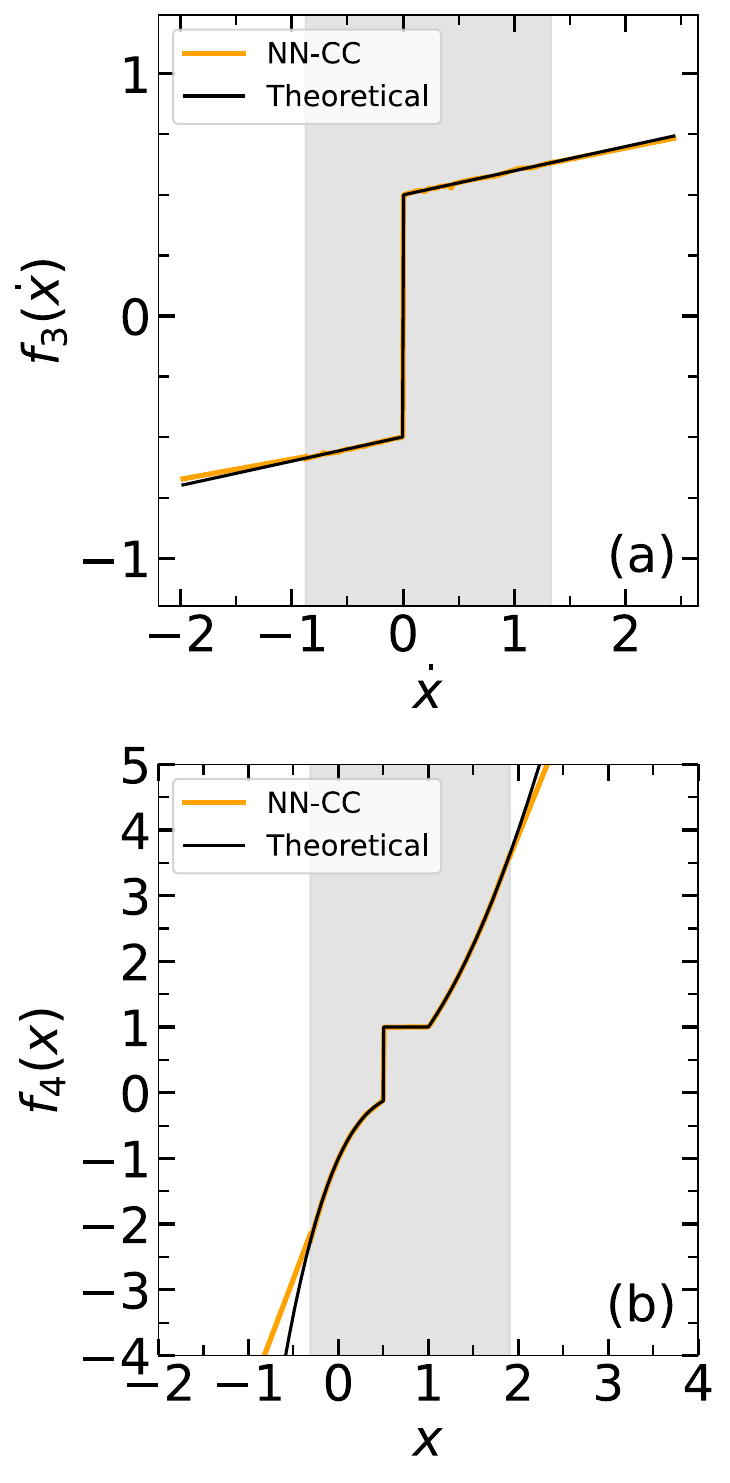}
    \hfill
    \includegraphics[width=0.32\linewidth]{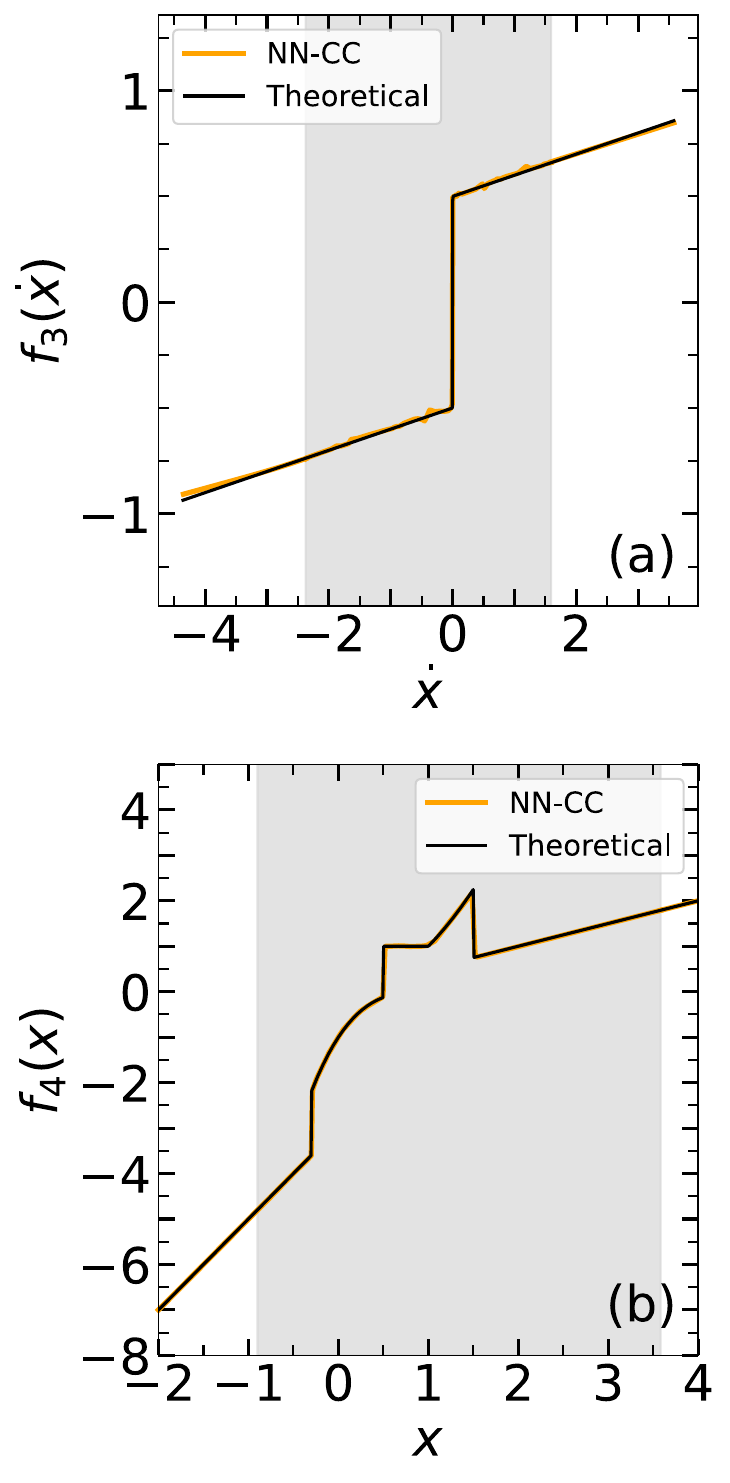}
    \caption{
    CCs identified using the NN-CC method for some representative velocity-dependent friction models: (a) $f_3(\dot{x})$ and (b) $f_4(x)$. Results are shown for three systems: Coulomb friction with backlash restoring function (left panels), Coulomb friction with piecewise restoring function 1 (middle panels), Coulomb friction with piecewise restoring function 2 (right panels).
    The shaded gray regions indicate the range of theoretical data used for identification.
    }
    \label{fig:friction-models_2}
\end{figure}





\section*{Acknowledgements}
The authors acknowledge computer time provided by CCT-Rosario Computational Center, member of the High Performance Computing National System (SNCAD, ME-Argentina).

\section*{Funding}
This work was partially supported by CONICET (Consejo Nacional de Investigaciones Científicas y Técnicas, Argentina) under Project PIP 1679.

\section*{Author contributions}
F.J.G. performed the simulations and wrote the initial draft of the manuscript. F.J.G and L.P.L contributed to the conceptual development, discussed the results, and participated in revising the manuscript. 

\section*{Competing interests}
The authors declare no competing interests.

\section*{Data availability}
Datasets generated during the current study are available from the corresponding author on reasonable request.

\bibliographystyle{unsrtnat}   
\bibliography{References}    

\begin{thebibliography}{78}
\providecommand{\natexlab}[1]{#1}
\providecommand{\url}[1]{\texttt{#1}}
\expandafter\ifx\csname urlstyle\endcsname\relax
  \providecommand{\doi}[1]{doi: #1}\else
  \providecommand{\doi}{doi: \begingroup \urlstyle{rm}\Url}\fi

\bibitem[Ljung(1999)]{Ljung1999}
L.~Ljung.
\newblock \emph{System Identification: Theory for the User}.
\newblock Prentice Hall, Upper Saddle River, NJ, second edition, 1999.
\newblock ISBN 0136566952; 9780136566953.

\bibitem[Billings(2013)]{Billings_2013}
Stephen~A Billings.
\newblock \emph{Nonlinear System Identification: NARMAX Methods in the Time,
  Frequency, and Spatio-Temporal Domains}.
\newblock John Wiley {\&} Sons, Ltd, jul 2013.
\newblock \doi{10.1002/9781118535561}.

\bibitem[Brunton et~al.(2016)Brunton, Proctor, and Kutz]{brunton2016}
Steven~L. Brunton, Joshua~L. Proctor, and J.~Nathan Kutz.
\newblock Discovering governing equations from data by sparse identification of
  nonlinear dynamical systems.
\newblock \emph{Proceedings of the National Academy of Sciences}, 113\penalty0
  (15):\penalty0 3932--3937, 2016.
\newblock \doi{10.1073/pnas.1517384113}.
\newblock URL \url{https://www.pnas.org/doi/abs/10.1073/pnas.1517384113}.

\bibitem[Egan et~al.(2024)Egan, Li, and Carvalho]{Egan2024}
Kevin Egan, Weizhen Li, and Rui Carvalho.
\newblock Automatically discovering ordinary differential equations from data
  with sparse regression.
\newblock \emph{Communications Physics}, 7\penalty0 (1), January 2024.
\newblock ISSN 2399-3650.
\newblock \doi{10.1038/s42005-023-01516-2}.
\newblock URL \url{http://dx.doi.org/10.1038/s42005-023-01516-2}.

\bibitem[Fan et~al.(2023)Fan, Bonilla, O'Kane, and Sisson]{Fan2023}
Xuhui Fan, Edwin~V. Bonilla, Terence O'Kane, and Scott~A Sisson.
\newblock Free-form variational inference for {G}aussian process state-space
  models.
\newblock In Andreas Krause, Emma Brunskill, Kyunghyun Cho, Barbara Engelhardt,
  Sivan Sabato, and Jonathan Scarlett, editors, \emph{Proceedings of the 40th
  International Conference on Machine Learning}, volume 202 of
  \emph{Proceedings of Machine Learning Research}, pages 9603--9622. PMLR,
  23--29 Jul 2023.
\newblock URL \url{https://proceedings.mlr.press/v202/fan23a.html}.

\bibitem[Korda and Mezi{\'c}(2016)]{Korda2016}
Milan Korda and Igor Mezi{\'c}.
\newblock Linear predictors for nonlinear dynamical systems: Koopman operator
  meets model predictive control.
\newblock \emph{Autom.}, 93:\penalty0 149--160, 2016.
\newblock URL \url{https://api.semanticscholar.org/CorpusID:49322864}.

\bibitem[Jin et~al.(2025)Jin, Hou, Ge, Gao, Yi, Li, Feng, and Zhong]{Jin2025}
Yuhong Jin, Lei Hou, Xiangdong Ge, Qiang Gao, Haiming Yi, Zhonggang Li, Yongzhi
  Feng, and Shun Zhong.
\newblock A novel data-driven modeling and efficient model predictive control
  framework for non-autonomous nonlinear systems based on the invertible
  koopman network.
\newblock \emph{Nonlinear Dynamics}, May 2025.
\newblock ISSN 1573-269X.
\newblock \doi{10.1007/s11071-025-11277-y}.
\newblock URL \url{http://dx.doi.org/10.1007/s11071-025-11277-y}.

\bibitem[Cranmer(2023)]{Cranmer2023PySR}
Miles Cranmer.
\newblock Interpretable machine learning for science with pysr and
  symbolicregression.jl, 2023.
\newblock URL \url{https://arxiv.org/abs/2305.01582}.

\bibitem[Schmidt and Lipson(2009)]{Schmidt2009}
Michael Schmidt and Hod Lipson.
\newblock Distilling free-form natural laws from experimental data.
\newblock \emph{Science}, 324\penalty0 (5923):\penalty0 81–85, April 2009.
\newblock ISSN 1095-9203.
\newblock \doi{10.1126/science.1165893}.
\newblock URL \url{http://dx.doi.org/10.1126/science.1165893}.

\bibitem[Makke and Chawla(2024)]{Makke2024}
Nour Makke and Sanjay Chawla.
\newblock Interpretable scientific discovery with symbolic regression: a
  review.
\newblock \emph{Artificial Intelligence Review}, 57\penalty0 (1), January 2024.
\newblock ISSN 1573-7462.
\newblock \doi{10.1007/s10462-023-10622-0}.
\newblock URL \url{http://dx.doi.org/10.1007/s10462-023-10622-0}.

\bibitem[Sjöberg et~al.(1994)Sjöberg, Hjalmarsson, and Ljung]{Sjoberg1994}
J.~Sjöberg, H.~Hjalmarsson, and L.~Ljung.
\newblock Neural networks in system identification.
\newblock \emph{IFAC Proceedings Volumes}, 27\penalty0 (8):\penalty0 359--382,
  1994.
\newblock ISSN 1474-6670.
\newblock \doi{https://doi.org/10.1016/S1474-6670(17)47737-8}.
\newblock URL
  \url{https://www.sciencedirect.com/science/article/pii/S1474667017477378}.
\newblock IFAC Symposium on System Identification (SYSID'94), Copenhagen,
  Denmark, 4-6 July.

\bibitem[Raissi et~al.(2019)Raissi, Perdikaris, and Karniadakis]{Raissi2019}
M.~Raissi, P.~Perdikaris, and G.E. Karniadakis.
\newblock Physics-informed neural networks: A deep learning framework for
  solving forward and inverse problems involving nonlinear partial differential
  equations.
\newblock \emph{Journal of Computational Physics}, 378:\penalty0 686--707,
  2019.
\newblock ISSN 0021-9991.
\newblock \doi{https://doi.org/10.1016/j.jcp.2018.10.045}.
\newblock URL
  \url{https://www.sciencedirect.com/science/article/pii/S0021999118307125}.

\bibitem[Faroughi et~al.(2024)Faroughi, Pawar, Fernandes, Raissi, Das,
  Kalantari, and Kourosh~Mahjour]{Faroughi2024}
Salah~A. Faroughi, Nikhil~M. Pawar, Célio Fernandes, Maziar Raissi, Subasish
  Das, Nima~K. Kalantari, and Seyed Kourosh~Mahjour.
\newblock Physics-guided, physics-informed, and physics-encoded neural networks
  and operators in scientific computing: Fluid and solid mechanics.
\newblock \emph{Journal of Computing and Information Science in Engineering},
  24\penalty0 (4):\penalty0 040802, 01 2024.
\newblock ISSN 1530-9827.
\newblock \doi{10.1115/1.4064449}.
\newblock URL \url{https://doi.org/10.1115/1.4064449}.

\bibitem[Brunton and Kutz(2022)]{Brunton_Kutz_2022}
Steven~L. Brunton and J.~Nathan Kutz.
\newblock \emph{Data-Driven Science and Engineering: Machine Learning,
  Dynamical Systems, and Control}.
\newblock Cambridge University Press, 2 edition, 2022.

\bibitem[Pillonetto et~al.(2025)Pillonetto, Aravkin, Gedon, Ljung, Ribeiro, and
  Schön]{Pilloneto2025}
Gianluigi Pillonetto, Aleksandr Aravkin, Daniel Gedon, Lennart Ljung,
  Antônio~H. Ribeiro, and Thomas~B. Schön.
\newblock Deep networks for system identification: A survey.
\newblock \emph{Automatica}, 171:\penalty0 111907, 2025.
\newblock ISSN 0005-1098.
\newblock \doi{https://doi.org/10.1016/j.automatica.2024.111907}.
\newblock URL
  \url{https://www.sciencedirect.com/science/article/pii/S0005109824004011}.

\bibitem[Goodfellow et~al.(2016)Goodfellow, Bengio, and
  Courville]{Goodfellow2016}
Ian Goodfellow, Yoshua Bengio, and Aaron Courville.
\newblock \emph{Deep Learning}.
\newblock MIT Press, 2016.
\newblock \url{http://www.deeplearningbook.org}.

\bibitem[Ayankoso and Olejnik(2023)]{Ayankoso2023}
Samuel Ayankoso and Paweł Olejnik.
\newblock Time-series machine learning techniques for modeling and
  identification of mechatronic systems with friction: A review and real
  application.
\newblock \emph{Electronics}, 12\penalty0 (17), 2023.
\newblock ISSN 2079-9292.
\newblock \doi{10.3390/electronics12173669}.
\newblock URL \url{https://www.mdpi.com/2079-9292/12/17/3669}.

\bibitem[Elaarabi et~al.(2025)Elaarabi, Borzacchiello, Bot, Guennec, and
  Comas-Cardona]{Elaarabi2025}
Mouad Elaarabi, Domenico Borzacchiello, Philippe~Le Bot, Yves L.~E. Guennec,
  and Sebastien Comas-Cardona.
\newblock Adaptive parameters identification for nonlinear dynamics using deep
  permutation invariant networks.
\newblock \emph{Mach. Learn.}, 114\penalty0 (1), January 2025.
\newblock ISSN 0885-6125.
\newblock \doi{10.1007/s10994-024-06732-7}.
\newblock URL \url{https://doi.org/10.1007/s10994-024-06732-7}.

\bibitem[Gonzalez(2023)]{Gonzalez2023}
Federico~J. Gonzalez.
\newblock Determination of the characteristic curves of a nonlinear first order
  system from fourier analysis.
\newblock \emph{Sci. Rep.}, 13\penalty0 (1):\penalty0 1955, February 2023.
\newblock \doi{10.1038/s41598-023-29151-5}.
\newblock URL \url{https://doi.org/10.1038/s41598-023-29151-5}.

\bibitem[Gonzalez(2024)]{Gonzalez2024}
Federico~J. Gonzalez.
\newblock System identification based on characteristic curves: a mathematical
  connection between power series and fourier analysis for first-order
  nonlinear systems.
\newblock \emph{Nonlinear Dynamics}, 112\penalty0 (18):\penalty0 16167–16197,
  July 2024.
\newblock ISSN 1573-269X.
\newblock \doi{10.1007/s11071-024-09890-4}.
\newblock URL \url{http://dx.doi.org/10.1007/s11071-024-09890-4}.

\bibitem[Warminski(2019)]{Warminski2019}
Jerzy Warminski.
\newblock Nonlinear dynamics of self-, parametric, and externally excited
  oscillator with time delay: van der pol versus rayleigh models.
\newblock \emph{Nonlinear Dynamics}, 99\penalty0 (1):\penalty0 35–56, June
  2019.
\newblock ISSN 1573-269X.
\newblock \doi{10.1007/s11071-019-05076-5}.
\newblock URL \url{http://dx.doi.org/10.1007/s11071-019-05076-5}.

\bibitem[Chua(1980)]{Chua1980}
L.~Chua.
\newblock Dynamic nonlinear networks: State-of-the-art.
\newblock \emph{IEEE Transactions on Circuits and Systems}, 27\penalty0
  (11):\penalty0 1059--1087, 1980.
\newblock \doi{10.1109/TCS.1980.1084745}.

\bibitem[Chua(2000)]{Chua2000}
L.~Chua.
\newblock \emph{Linear and Non Linear Circuits}.
\newblock McGraw-Hill Education, 2000.

\bibitem[Winters and Stark(1987)]{Winters1987}
J.~M. Winters and L.~Stark.
\newblock Muscle models: What is gained and what is lost by varying model
  complexity.
\newblock \emph{Biological Cybernetics}, 55\penalty0 (6):\penalty0 403–420,
  March 1987.
\newblock ISSN 1432-0770.
\newblock \doi{10.1007/bf00318375}.
\newblock URL \url{http://dx.doi.org/10.1007/BF00318375}.

\bibitem[Özcan et~al.(2013)Özcan, Sönmez, and Güvenç]{Ozcan2013}
Dinçer Özcan, Ümit Sönmez, and Levent Güvenç.
\newblock Optimisation of the nonlinear suspension characteristics of a light
  commercial vehicle.
\newblock \emph{International Journal of Vehicular Technology}, 2013\penalty0
  (1):\penalty0 562424, 2013.
\newblock \doi{https://doi.org/10.1155/2013/562424}.
\newblock URL
  \url{https://onlinelibrary.wiley.com/doi/abs/10.1155/2013/562424}.

\bibitem[van~der Pol(1926)]{vanderPol1926}
Balth. van~der Pol.
\newblock Lxxxviii.on “relaxation-oscillations”.
\newblock \emph{The London, Edinburgh, and Dublin Philosophical Magazine and
  Journal of Science}, 2\penalty0 (11):\penalty0 978–992, November 1926.
\newblock ISSN 1941-5990.
\newblock \doi{10.1080/14786442608564127}.
\newblock URL \url{http://dx.doi.org/10.1080/14786442608564127}.

\bibitem[Farkas(1994)]{Farkas1994}
Miklós Farkas.
\newblock \emph{Periodic Motions}.
\newblock Springer New York, 1994.
\newblock ISBN 9781475742114.
\newblock \doi{10.1007/978-1-4757-4211-4}.
\newblock URL \url{http://dx.doi.org/10.1007/978-1-4757-4211-4}.

\bibitem[Wu et~al.(2020)Wu, Chen, and Wang]{Wu2020}
Yanchi Wu, Xinzhong Chen, and Yunfei Wang.
\newblock Identification of nonlinear aerodynamic damping of wind-excited
  flexible structures by curve-fitting non-gaussian response probability
  density function.
\newblock \emph{Journal of Wind Engineering and Industrial Aerodynamics},
  206:\penalty0 104311, 2020.
\newblock ISSN 0167-6105.
\newblock \doi{https://doi.org/10.1016/j.jweia.2020.104311}.
\newblock URL
  \url{https://www.sciencedirect.com/science/article/pii/S016761052030221X}.

\bibitem[Holmes(1979)]{Holmes1979}
P.~Holmes.
\newblock A nonlinear oscillator with a strange attractor.
\newblock \emph{Philosophical Transactions of the Royal Society of London.
  Series A, Mathematical and Physical Sciences}, 292\penalty0 (1394):\penalty0
  419--448, 1979.
\newblock ISSN 00804614.
\newblock URL \url{http://www.jstor.org/stable/75173}.

\bibitem[Burra and Zanolin(2025)]{Burra2025}
Lakshmi Burra and Fabio Zanolin.
\newblock \emph{The Duffing Equation: Periodic Solutions and Chaotic Dynamics}.
\newblock Springer Nature Singapore, 2025.
\newblock ISBN 9789819783014.
\newblock \doi{10.1007/978-981-97-8301-4}.
\newblock URL \url{http://dx.doi.org/10.1007/978-981-97-8301-4}.

\bibitem[Story and Titze(1995)]{Story1995}
Brad~H. Story and Ingo~R. Titze.
\newblock Voice simulation with a body-cover model of the vocal folds.
\newblock \emph{The Journal of the Acoustical Society of America}, 97\penalty0
  (2):\penalty0 1249–1260, February 1995.
\newblock ISSN 1520-8524.
\newblock \doi{10.1121/1.412234}.
\newblock URL \url{http://dx.doi.org/10.1121/1.412234}.

\bibitem[Cataldo and Soize(2021)]{Cataldo2021}
E.~Cataldo and C.~Soize.
\newblock A stochastic model of voice generation and the corresponding solution
  for the inverse problem using artificial neural network for case with
  pathology in the vocal folds.
\newblock \emph{Biomedical Signal Processing and Control}, 68:\penalty0 102623,
  2021.
\newblock ISSN 1746-8094.
\newblock \doi{https://doi.org/10.1016/j.bspc.2021.102623}.
\newblock URL
  \url{https://www.sciencedirect.com/science/article/pii/S1746809421002202}.

\bibitem[Zhao and Singh(2023)]{Zhao2023}
Wayne Zhao and Rita Singh.
\newblock Deriving vocal fold oscillation information from recorded voice
  signals using models of phonation.
\newblock \emph{Entropy}, 25\penalty0 (7):\penalty0 1039, July 2023.
\newblock ISSN 1099-4300.
\newblock \doi{10.3390/e25071039}.
\newblock URL \url{http://dx.doi.org/10.3390/e25071039}.

\bibitem[Bikdash et~al.(1994)Bikdash, Balachandran, and Navfeh]{Bikdash1994}
M.~Bikdash, B.~Balachandran, and A.~Navfeh.
\newblock Melnikov analysis for a ship with a general roll-damping model.
\newblock \emph{Nonlinear Dynamics}, 6\penalty0 (1):\penalty0 101–124, July
  1994.
\newblock ISSN 1573-269X.
\newblock \doi{10.1007/bf00045435}.
\newblock URL \url{http://dx.doi.org/10.1007/BF00045435}.

\bibitem[Mahfouz(2004)]{Mahfouz2004}
Ayman~B. Mahfouz.
\newblock Identification of the nonlinear ship rolling motion equation using
  the measured response at sea.
\newblock \emph{Ocean Engineering}, 31\penalty0 (17):\penalty0 2139--2156,
  2004.
\newblock ISSN 0029-8018.
\newblock \doi{https://doi.org/10.1016/j.oceaneng.2004.06.001}.
\newblock URL
  \url{https://www.sciencedirect.com/science/article/pii/S0029801804001295}.

\bibitem[Takami et~al.(2024)Takami, {Dam Nielsen}, {Juncher Jensen}, Maki,
  Matsui, and Komoriyama]{Takami2024}
Tomoki Takami, Ulrik {Dam Nielsen}, Jørgen {Juncher Jensen}, Atsuo Maki,
  Sadaoki Matsui, and Yusuke Komoriyama.
\newblock Onboard identification of stability parameters including nonlinear
  roll damping via phase-resolved wave estimation using measured ship
  responses.
\newblock \emph{Mechanical Systems and Signal Processing}, 210:\penalty0
  111166, 2024.
\newblock ISSN 0888-3270.
\newblock \doi{https://doi.org/10.1016/j.ymssp.2024.111166}.
\newblock URL
  \url{https://www.sciencedirect.com/science/article/pii/S0888327024000645}.

\bibitem[Chuang et~al.(2010)Chuang, Lee, Chang, and Hu]{Chuang2010}
Wan-Chun Chuang, Hsin-Li Lee, Pei-Zen Chang, and Yuh-Chung Hu.
\newblock Review on the modeling of electrostatic mems.
\newblock \emph{Sensors}, 10\penalty0 (6):\penalty0 6149--6171, 2010.
\newblock ISSN 1424-8220.
\newblock \doi{10.3390/s100606149}.
\newblock URL \url{https://www.mdpi.com/1424-8220/10/6/6149}.

\bibitem[Shaw and Holmes(1983)]{Shaw1983}
S.W. Shaw and P.J. Holmes.
\newblock A periodically forced piecewise linear oscillator.
\newblock \emph{Journal of Sound and Vibration}, 90\penalty0 (1):\penalty0
  129--155, 1983.
\newblock ISSN 0022-460X.
\newblock \doi{https://doi.org/10.1016/0022-460X(83)90407-8}.
\newblock URL
  \url{https://www.sciencedirect.com/science/article/pii/0022460X83904078}.

\bibitem[Shaw(1985)]{Shaw1985}
S.~W. Shaw.
\newblock The dynamics of a harmonically excited system having rigid amplitude
  constraints, part 1: Subharmonic motions and local bifurcations.
\newblock \emph{Journal of Applied Mechanics}, 52\penalty0 (2):\penalty0
  453–458, June 1985.
\newblock ISSN 1528-9036.
\newblock \doi{10.1115/1.3169068}.
\newblock URL \url{http://dx.doi.org/10.1115/1.3169068}.

\bibitem[Feng(2018)]{Feng2018}
Jinqian Feng.
\newblock Chaos controls of a duffing system with impacts.
\newblock \emph{AIP Advances}, 8\penalty0 (4):\penalty0 045303, 04 2018.
\newblock ISSN 2158-3226.
\newblock \doi{10.1063/1.5021965}.
\newblock URL \url{https://doi.org/10.1063/1.5021965}.

\bibitem[FitzHugh(1961)]{Fitzhugh1961}
Richard FitzHugh.
\newblock Impulses and physiological states in theoretical models of nerve
  membrane.
\newblock \emph{Biophysical Journal}, 1\penalty0 (6):\penalty0 445--466, 1961.
\newblock ISSN 0006-3495.
\newblock \doi{https://doi.org/10.1016/S0006-3495(61)86902-6}.
\newblock URL
  \url{https://www.sciencedirect.com/science/article/pii/S0006349561869026}.

\bibitem[Gerstner and Kistler(2002)]{Gerstner2002}
Wulfram Gerstner and Werner~M. Kistler.
\newblock \emph{Spiking Neuron Models: Single Neurons, Populations,
  Plasticity}.
\newblock Cambridge University Press, August 2002.
\newblock ISBN 9780511815706.
\newblock \doi{10.1017/cbo9780511815706}.
\newblock URL \url{http://dx.doi.org/10.1017/CBO9780511815706}.

\bibitem[Rudi et~al.(2020)Rudi, Bessac, and Lenzi]{Rudi2020}
Johann Rudi, Julie Bessac, and Amanda Lenzi.
\newblock Parameter estimation with dense and convolutional neural networks
  applied to the fitzhugh-nagumo ode.
\newblock In \emph{Mathematical and Scientific Machine Learning}, 2020.
\newblock URL \url{https://api.semanticscholar.org/CorpusID:229156866}.

\bibitem[Nayfeh and Mook(1979)]{nayfeh1979nonlinear}
Ali~H Nayfeh and Dean~T Mook.
\newblock \emph{Nonlinear oscillations}.
\newblock John Wiley \& Sons, New York, NY, 1979.

\bibitem[Senturia(2001)]{Senturia2001}
Stephen~D. Senturia.
\newblock \emph{Microsystem Design}.
\newblock Springer US, 2001.
\newblock ISBN 9780306476013.
\newblock \doi{10.1007/b117574}.
\newblock URL \url{http://dx.doi.org/10.1007/b117574}.

\bibitem[Lifshitz and Cross(2008)]{Lifshitz2008}
Ron Lifshitz and M.~C. Cross.
\newblock \emph{Nonlinear Dynamics of Nanomechanical and Micromechanical
  Resonators}, chapter~1, pages 1--52.
\newblock John Wiley \& Sons, Ltd, 2008.
\newblock ISBN 9783527626359.
\newblock \doi{https://doi.org/10.1002/9783527626359.ch1}.
\newblock URL
  \url{https://onlinelibrary.wiley.com/doi/abs/10.1002/9783527626359.ch1}.

\bibitem[Younis(2011)]{Younis2011}
Mohammad~I. Younis.
\newblock \emph{MEMS Linear and Nonlinear Statics and Dynamics}.
\newblock Springer US, 2011.
\newblock ISBN 9781441960207.
\newblock \doi{10.1007/978-1-4419-6020-7}.
\newblock URL \url{http://dx.doi.org/10.1007/978-1-4419-6020-7}.

\bibitem[Olsson et~al.(1998)Olsson, Åström, {Canudas de Wit}, Gäfvert, and
  Lischinsky]{Olsson1998}
H.~Olsson, K.J. Åström, C.~{Canudas de Wit}, M.~Gäfvert, and P.~Lischinsky.
\newblock Friction models and friction compensation.
\newblock \emph{European Journal of Control}, 4\penalty0 (3):\penalty0
  176--195, 1998.
\newblock ISSN 0947-3580.
\newblock \doi{https://doi.org/10.1016/S0947-3580(98)70113-X}.
\newblock URL
  \url{https://www.sciencedirect.com/science/article/pii/S094735809870113X}.

\bibitem[Rill and Schuderer(2023)]{Rill2023}
Georg Rill and Matthias Schuderer.
\newblock A second-order dynamic friction model compared to commercial
  stick–slip models.
\newblock \emph{Modelling}, 4\penalty0 (3):\penalty0 366--381, 2023.
\newblock ISSN 2673-3951.
\newblock \doi{10.3390/modelling4030021}.
\newblock URL \url{https://www.mdpi.com/2673-3951/4/3/21}.

\bibitem[González-Carbajal et~al.(2024)González-Carbajal, García-Vallejo,
  Domínguez, and Freire]{GonzalezCarbajal2024}
J.~González-Carbajal, D.~García-Vallejo, J.~Domínguez, and E.~Freire.
\newblock The role of dynamic friction in the appearance of periodic
  oscillations in mechanical systems.
\newblock \emph{Nonlinear Dynamics}, 112\penalty0 (24):\penalty0 21587–21603,
  September 2024.
\newblock ISSN 1573-269X.
\newblock \doi{10.1007/s11071-024-10162-4}.
\newblock URL \url{http://dx.doi.org/10.1007/s11071-024-10162-4}.

\bibitem[Dieterich(1979{\natexlab{a}})]{Dieterich1979part1}
James~H. Dieterich.
\newblock Modeling of rock friction: 1. experimental results and constitutive
  equations.
\newblock \emph{Journal of Geophysical Research: Solid Earth}, 84\penalty0
  (B5):\penalty0 2161–2168, May 1979{\natexlab{a}}.
\newblock ISSN 0148-0227.
\newblock \doi{10.1029/jb084ib05p02161}.
\newblock URL \url{http://dx.doi.org/10.1029/JB084iB05p02161}.

\bibitem[Dieterich(1979{\natexlab{b}})]{Dieterich1979part2}
James~H. Dieterich.
\newblock Modeling of rock friction: 2. simulation of preseismic slip.
\newblock \emph{Journal of Geophysical Research: Solid Earth}, 84\penalty0
  (B5):\penalty0 2169–2175, May 1979{\natexlab{b}}.
\newblock ISSN 0148-0227.
\newblock \doi{10.1029/jb084ib05p02169}.
\newblock URL \url{http://dx.doi.org/10.1029/JB084iB05p02169}.

\bibitem[Stribeck(1902)]{Stribeck1902}
R.~Stribeck.
\newblock Die wesentlichen eigenschaften der gleit- und rollenlager.
\newblock \emph{Zeitschrift des Vereins Deutscher Ingenieure}, 46:\penalty0
  1341--1348, 1902.

\bibitem[Canudas~de Wit et~al.(1995)Canudas~de Wit, Olsson, Astrom, and
  Lischinsky]{CanudasdeWit1995}
C.~Canudas~de Wit, H.~Olsson, K.J. Astrom, and P.~Lischinsky.
\newblock A new model for control of systems with friction.
\newblock \emph{IEEE Transactions on Automatic Control}, 40\penalty0
  (3):\penalty0 419–425, March 1995.
\newblock ISSN 0018-9286.
\newblock \doi{10.1109/9.376053}.
\newblock URL \url{http://dx.doi.org/10.1109/9.376053}.

\bibitem[Braza(2020)]{Braza2020}
Peter~A. Braza.
\newblock The interplay of damping and amplitude in the nonlinear pendulum.
\newblock \emph{American Journal of Physics}, 88\penalty0 (5):\penalty0
  379--384, 05 2020.
\newblock ISSN 0002-9505.
\newblock \doi{10.1119/10.0000630}.
\newblock URL \url{https://doi.org/10.1119/10.0000630}.

\bibitem[Pal et~al.(2023)Pal, Ray, Nag~Chowdhury, and Ghosh]{Pal2023}
Tapas~Kumar Pal, Arnob Ray, Sayantan Nag~Chowdhury, and Dibakar Ghosh.
\newblock Extreme rotational events in a forced-damped nonlinear pendulum.
\newblock \emph{Chaos: An Interdisciplinary Journal of Nonlinear Science},
  33\penalty0 (6):\penalty0 063134, 06 2023.
\newblock ISSN 1054-1500.
\newblock \doi{10.1063/5.0152699}.
\newblock URL \url{https://doi.org/10.1063/5.0152699}.

\bibitem[Ji(2004)]{Ji2004}
J.C. Ji.
\newblock Dynamics of a piecewise linear system subjected to a saturation
  constraint.
\newblock \emph{Journal of Sound and Vibration}, 271\penalty0 (3):\penalty0
  905--920, 2004.
\newblock ISSN 0022-460X.
\newblock \doi{https://doi.org/10.1016/S0022-460X(03)00759-4}.
\newblock URL
  \url{https://www.sciencedirect.com/science/article/pii/S0022460X03007594}.

\bibitem[Bustamante et~al.(2025)Bustamante, Rajagopal, and
  Rodriguez]{Bustamante2025}
R~Bustamante, K~R Rajagopal, and C~Rodriguez.
\newblock Piecewise linear constitutive relations for stretch-limited elastic
  strings.
\newblock \emph{IMA Journal of Applied Mathematics}, April 2025.
\newblock ISSN 1464-3634.
\newblock \doi{10.1093/imamat/hxaf010}.
\newblock URL \url{http://dx.doi.org/10.1093/imamat/hxaf010}.

\bibitem[Slotine and Li(1991)]{slotine1991}
J.~E. Slotine and Weiping Li.
\newblock \emph{Applied nonlinear control}.
\newblock Prentice Hall, Englewood Cliffs, NJ, 1991.

\bibitem[Nordin and Gutman(2002)]{Nordin2002}
Mattias Nordin and Per-Olof Gutman.
\newblock Controlling mechanical systems with backlash—a survey.
\newblock \emph{Automatica}, 38\penalty0 (10):\penalty0 1633--1649, 2002.
\newblock ISSN 0005-1098.
\newblock \doi{https://doi.org/10.1016/S0005-1098(02)00047-X}.
\newblock URL
  \url{https://www.sciencedirect.com/science/article/pii/S000510980200047X}.

\bibitem[Karimov et~al.(2021)Karimov, Kopets, Nepomuceno, and
  Butusov]{karimov2021}
Artur~I. Karimov, Ekaterina Kopets, Erivelton~G. Nepomuceno, and Denis Butusov.
\newblock Integrate-and-differentiate approach to nonlinear system
  identification.
\newblock \emph{Mathematics}, 9\penalty0 (23), 2021.
\newblock ISSN 2227-7390.
\newblock \doi{10.3390/math9232999}.
\newblock URL \url{https://www.mdpi.com/2227-7390/9/23/2999}.

\bibitem[Tikhonov et~al.(1995)Tikhonov, Goncharsky, Stepanov, and
  Yagola]{Tikhonov1995}
A.~N. Tikhonov, A.~V. Goncharsky, V.~V. Stepanov, and A.~G. Yagola.
\newblock \emph{Numerical Methods for the Solution of Ill-Posed Problems}.
\newblock Springer Netherlands, 1995.
\newblock ISBN 9789401584807.
\newblock \doi{10.1007/978-94-015-8480-7}.
\newblock URL \url{http://dx.doi.org/10.1007/978-94-015-8480-7}.

\bibitem[Freeden and Zuhair~Nashed(2018)]{Freeden2018}
Willi Freeden and M.~Zuhair~Nashed.
\newblock \emph{Ill-Posed Problems: Operator Methodologies of Resolution and
  Regularization}, pages 201--314.
\newblock Springer International Publishing, Cham, 2018.
\newblock ISBN 978-3-319-57181-2.
\newblock \doi{10.1007/978-3-319-57181-2_3}.
\newblock URL \url{https://doi.org/10.1007/978-3-319-57181-2_3}.

\bibitem[Kaptanoglu et~al.(2021)Kaptanoglu, Morgan, Hansen, and
  Brunton]{Kaptanoglu2021}
Alan~A. Kaptanoglu, Kyle~D. Morgan, Chris~J. Hansen, and Steven~L. Brunton.
\newblock Physics-constrained, low-dimensional models for magnetohydrodynamics:
  First-principles and data-driven approaches.
\newblock \emph{Phys. Rev. E}, 104:\penalty0 015206, Jul 2021.
\newblock \doi{10.1103/PhysRevE.104.015206}.
\newblock URL \url{https://link.aps.org/doi/10.1103/PhysRevE.104.015206}.

\bibitem[Kaheman et~al.(2022)Kaheman, Brunton, and Nathan~Kutz]{Kaheman2022}
Kadierdan Kaheman, Steven~L Brunton, and J~Nathan~Kutz.
\newblock Automatic differentiation to simultaneously identify nonlinear
  dynamics and extract noise probability distributions from data.
\newblock \emph{Machine Learning: Science and Technology}, 3\penalty0
  (1):\penalty0 015031, March 2022.
\newblock ISSN 2632-2153.
\newblock \doi{10.1088/2632-2153/ac567a}.
\newblock URL \url{http://dx.doi.org/10.1088/2632-2153/ac567a}.

\bibitem[Strebel(2025)]{Strebel2025}
Oliver Strebel.
\newblock Numerical differentiation by integrated series expansion (ndbise) in
  the context of ordinary differential equation estimation problems, Apr 2025.
\newblock URL \url{osf.io/k9dy6_v1}.

\bibitem[Hindmarsh(1983)]{Hindmarsh1983}
Alan~C. Hindmarsh.
\newblock {ODEPACK, A Systematized Collection of ODE Solvers}.
\newblock In R.~S. Stepleman et~al., editors, \emph{Scientific Computing},
  pages 55--64, Amsterdam, 1983. North-Holland.
\newblock IMACS Transactions on Scientific Computation, Vol. 1.

\bibitem[Petzold(1983)]{Petzold1983}
Linda Petzold.
\newblock Automatic selection of methods for solving stiff and nonstiff systems
  of ordinary differential equations.
\newblock \emph{SIAM Journal on Scientific and Statistical Computing},
  4\penalty0 (1):\penalty0 136--148, 1983.
\newblock \doi{10.1137/0904010}.
\newblock URL \url{https://doi.org/10.1137/0904010}.

\bibitem[Press et~al.(2007)Press, Teukolsky, Vetterling, and
  Flannery]{William2007numericalrecipes}
William~H. Press, Saul~A. Teukolsky, William~T. Vetterling, and Brian~P.
  Flannery.
\newblock \emph{Numerical Recipes 3rd Edition: The Art of Scientific
  Computing}.
\newblock Cambridge University Press, USA, 3 edition, 2007.
\newblock ISBN 0521880688.

\bibitem[Cybenko(1989)]{Cybenko1989}
G.~Cybenko.
\newblock Approximation by superpositions of a sigmoidal function.
\newblock \emph{Mathematics of Control, Signals, and Systems}, 2\penalty0
  (4):\penalty0 303–314, December 1989.
\newblock ISSN 1435-568X.
\newblock \doi{10.1007/bf02551274}.
\newblock URL \url{http://dx.doi.org/10.1007/BF02551274}.

\bibitem[Pinkus(1999)]{Pinkus1999}
Allan Pinkus.
\newblock Approximation theory of the mlp model in neural networks.
\newblock \emph{Acta Numerica}, 8:\penalty0 143–195, 1999.
\newblock \doi{10.1017/S0962492900002919}.

\bibitem[Hornik et~al.(1989)Hornik, Stinchcombe, and White]{Hornik1989}
Kurt Hornik, Maxwell Stinchcombe, and Halbert White.
\newblock Multilayer feedforward networks are universal approximators.
\newblock \emph{Neural Networks}, 2\penalty0 (5):\penalty0 359--366, 1989.
\newblock ISSN 0893-6080.
\newblock \doi{https://doi.org/10.1016/0893-6080(89)90020-8}.
\newblock URL
  \url{https://www.sciencedirect.com/science/article/pii/0893608089900208}.

\bibitem[Chandra et~al.(2024)Chandra, Kapoor, Curti, Tiels, and
  Lomonova]{Chandra2024deep}
Abhishek Chandra, Taniya Kapoor, Mitrofan Curti, Koen Tiels, and Elena~A.
  Lomonova.
\newblock Characterizing nonlinear piezoelectric dynamics through deep neural
  operator learning.
\newblock \emph{Applied Physics Letters}, 125\penalty0 (26):\penalty0 262902,
  12 2024.
\newblock ISSN 0003-6951.
\newblock \doi{10.1063/5.0239160}.
\newblock URL \url{https://doi.org/10.1063/5.0239160}.

\bibitem[Chandra et~al.(2025)Chandra, Kapoor, Daniels, Curti, Tiels,
  Tartakovsky, and Lomonova]{Chandra2025HistRNN}
Abhishek Chandra, Taniya Kapoor, Bram Daniels, Mitrofan Curti, Koen Tiels,
  Daniel~M. Tartakovsky, and Elena~A. Lomonova.
\newblock Generalizable models of magnetic hysteresis via physics-aware
  recurrent neural networks.
\newblock \emph{Computer Physics Communications}, 314:\penalty0 109650, 2025.
\newblock ISSN 0010-4655.
\newblock \doi{https://doi.org/10.1016/j.cpc.2025.109650}.
\newblock URL
  \url{https://www.sciencedirect.com/science/article/pii/S0010465525001523}.

\bibitem[Lathourakis and Cicirello(2024)]{Lathourakis2024}
Christos Lathourakis and Alice Cicirello.
\newblock Physics enhanced sparse identification of dynamical systems with
  discontinuous nonlinearities.
\newblock \emph{Nonlinear Dynamics}, 112\penalty0 (13):\penalty0 11237–11264,
  May 2024.
\newblock ISSN 1573-269X.
\newblock \doi{10.1007/s11071-024-09652-2}.
\newblock URL \url{http://dx.doi.org/10.1007/s11071-024-09652-2}.

\bibitem[Karniadakis et~al.(2021)Karniadakis, Kevrekidis, Lu, Perdikaris, Wang,
  and Yang]{Karniadakis2021}
George~Em Karniadakis, Ioannis~G. Kevrekidis, Lu~Lu, Paris Perdikaris, Sifan
  Wang, and Liu Yang.
\newblock Physics-informed machine learning.
\newblock \emph{Nature Reviews Physics}, 3\penalty0 (6):\penalty0 422–440,
  May 2021.
\newblock ISSN 2522-5820.
\newblock \doi{10.1038/s42254-021-00314-5}.
\newblock URL \url{http://dx.doi.org/10.1038/s42254-021-00314-5}.

\bibitem[Sun et~al.(2023)Sun, Cai, Peng, Cheng, Wang, and Zhang]{Sun2023}
B.~Sun, Q.~Y. Cai, Z.~K. Peng, C.~M. Cheng, F.~Wang, and H.~Z. Zhang.
\newblock Variable selection and identification of high-dimensional
  nonparametric nonlinear systems by directional regression.
\newblock \emph{Nonlinear Dynamics}, 111\penalty0 (13):\penalty0 12101–12112,
  May 2023.
\newblock ISSN 1573-269X.
\newblock \doi{10.1007/s11071-023-08488-6}.
\newblock URL \url{http://dx.doi.org/10.1007/s11071-023-08488-6}.

\bibitem[Jordan and Smith(2007)]{Jordan2007}
D.~W. Jordan and P.~Smith.
\newblock \emph{Nonlinear Ordinary Differential Equations: An introduction for
  Scientists and Engineers}.
\newblock Oxford University Press, Oxford, August 2007.
\newblock ISBN 9781383034752.
\newblock \doi{10.1093/oso/9780199208241.001.0001}.
\newblock URL \url{http://dx.doi.org/10.1093/oso/9780199208241.001.0001}.

\end{thebibliography}

\end{document}